%% file: main.tex
\pdfoutput=1
\documentclass[10pt,twocolumn,letterpaper]{article}

\usepackage[pagenumbers]{cvpr} 

\input{preamble}

\definecolor{cvprblue}{rgb}{0.21,0.49,0.74}
\usepackage[pagebackref,breaklinks,colorlinks,allcolors=cvprblue]{hyperref}

\title{Which Computations Drive Visual-to-Text Transfer?\\A Causal Ledger for Vision-Language Models}

\author{
Zuming Mei$^{1,*}$ \quad Ruixin Liu$^{1,*}$ \quad Daliang Xu$^{2,\dagger}$ \quad Qi Geng$^{3}$ \quad Xin Li$^{3}$ \quad Xuanzhe Liu$^{1,\dagger}$\\[2pt]
{\normalsize $^{1}$Peking University \qquad $^{2}$Beijing University of Posts and Telecommunications}\\
{\normalsize $^{3}$Inspur Smart City Technology Co., Ltd.}\\[2pt]
{\small $^{*}$Equal contribution \qquad $^{\dagger}$Corresponding authors}
}

\begin{document}

\twocolumn[{%
\renewcommand\twocolumn[1][]{#1}%
\maketitle
\vspace{-2.4em}
\begin{center}
  \centering
  \captionsetup{type=figure}
  \includegraphics[width=0.95\textwidth]{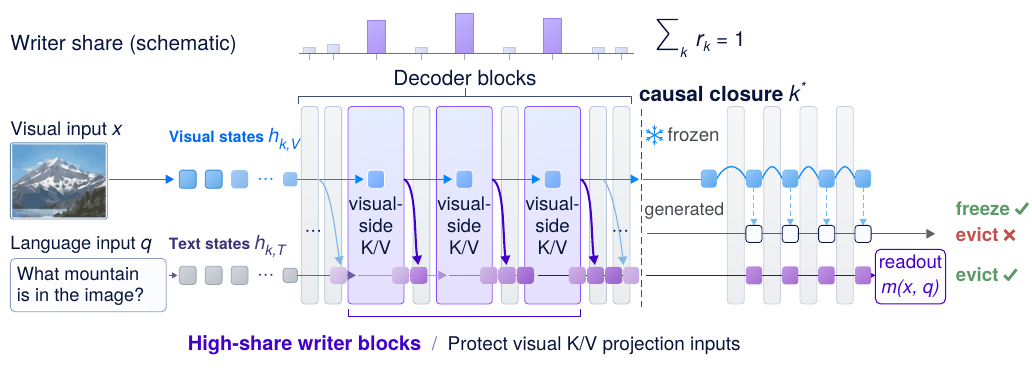}
  \caption{Schematic of visual-to-text transfer. Text queries read visual rows (blue) via the K/V
  projections of specific writer blocks (shaded), transferring answer-relevant evidence into text
  states (purple arrows). \textit{Writer shares} $r_k$ quantify each block's fractional contribution
  to the total answer effect (summing to 1). \textit{Causal closure} $k^*$ marks the boundary where
  residual visual dependence becomes persistently negligible. After $k^*$, later blocks no longer
  need to update the visual rows, which can therefore be frozen. The readout $m(x,q)$ of a short
  answer at the end of the prompt does not read them either, so they can also be evicted. Tokens
  generated in multi-token reasoning still read the frozen rows (dashed arrows), and only freezing
  keeps them readable.}
  \label{fig:intro-transfer}
\end{center}%
\vspace{-0.8em}
}]

\input{sections/0_abstract}
\input{sections/introduction}

\input{sections/related_work}
\input{sections/method_setup}
\input{sections/method_validity}
\input{sections/method_predictions}
\input{sections/experiments_setup}
\input{sections/experiments_compression_signal}
\input{sections/exp_ablations}
\input{sections/exp_limits}
\input{sections/conclusion}

{
    \small
    \bibliographystyle{ieeenat_fullname}
    \bibliography{main}
}

\clearpage
\onecolumn
\begin{center}
  \Large\textbf{\thetitle}\\
  \vspace{0.5em}Supplementary Material\\
  \vspace{1.0em}
\end{center}
\appendix
\input{sections/appendix}

\end{document}

%% file: preamble.tex
\input{math_commands.tex}

\usepackage{url}
\usepackage{longtable}
\usepackage{float}
\usepackage{enumitem}
\usepackage{multirow}
\usepackage{colortbl}
\usepackage{pifont}
\newcommand{\cmark}{\ding{51}}
\newcommand{\xmark}{\ding{55}}
\usepackage{tikz}
\usetikzlibrary{arrows.meta}
\graphicspath{{figures/}}

%% file: math_commands.tex
\usepackage{amsmath,amsfonts,bm}

\def\eqref#1{equation~\ref{#1}}

\def\1{\bm{1}}

\DeclareMathAlphabet{\mathsfit}{\encodingdefault}{\sfdefault}{m}{sl}
\SetMathAlphabet{\mathsfit}{bold}{\encodingdefault}{\sfdefault}{bx}{n}



%% file: sections/0_abstract.tex
\begin{abstract}
Vision-language models can incorporate visual evidence into text states, but how much
each decoder block contributes to the answer-relevant visual effect remains difficult
to quantify. We introduce reciprocal nested interventions that decompose a common
image-conditioned answer effect into ordered block contributions. By replacing visual
rows at successive decoder boundaries while retaining text states, we measure how much
of the visual effect has entered text states by each depth. These contributions sum
exactly to the total effect, yielding a layerwise ledger of cumulative transfer
and residual visual dependence. Transfer is highly concentrated: just two writer blocks carry
$83\%$ and $71\%$ of the total visual effect in Qwen3.5-0.8B and Qwen3.5-4B, respectively, and three carry
$72\%$ in LLaVA-1.5-7B. The ledger makes two predictions relevant to compression. Its causal closure predicts how
early visual tokens can be evicted for short answers, while writer share prioritizes which
blocks' $K/V$ inputs to protect in BF16 under quantization. During generation, however, eviction at
closure costs $18$~pp on MMStar mathematics, because generated tokens still read the visual
rows in late blocks. Freezing the rows at closure, which stops their updates but keeps them
readable, recovers most of this loss, so closure marks where computation on visual rows can
stop, not where they can be discarded.
\end{abstract}

%% file: sections/introduction.tex
\section{Introduction}
\label{sec:introduction}

Understanding how visual information flows through vision-language models (VLMs) is central to interpreting their reasoning. 
Prior mechanistic studies have identified specific pathways for this flow, showing that visual evidence transfers to text 
states, such as question-token representations \citep{zhang2025crossmodal} or prediction positions \citep{neo2025interpreting}, during 
generation. Building on this, recent inference techniques like VTW have demonstrated that visual tokens can be entirely dropped 
at certain depths with little performance degradation in evaluated settings \citep{lin2025vtw}. Counterfactual patching and attention knockout likewise find that image tokens lose influence after the middle layers \citep{kang2026linear, kim2026mapflow}. However, a critical gap remains: knowing \textit{when} 
visual tokens become dispensable does not explain \textit{how} this visual-to-text transfer accumulated across 
preceding layers. While it is natural to hypothesize that earlier computations established this effect, current 
withdrawal criteria fail to quantify how much each specific block contributes to the final answer, or how this transfer 
dictates precision sensitivity.

Existing interpretability tools struggle to fully answer this question. While attention analysis characterizes visual access 
\citep{chen2024fastv, zhang2025sparsevlm, xing2025pyramiddrop}, it does not in general reflect true causal contributions 
\citep{jain2019attention, wiegreffe2019attention}. Conversely, causal tracing \citep{basu2024storage}, counterfactual patching 
\citep{salazar2026pathways, kang2026linear}, and pathway interventions \citep{zhang2025crossmodal, neo2025interpreting, kim2026mapflow} 
identify important components and connections. Yet, their intervention effects at different locations rarely sum up to the total 
visual effect due to internal model interactions and compensation. What is missing is a unified accounting framework—one where 
cumulative transfer, per-layer contribution, and residual visual dependence are all measured against the same answer-level effect. 
To address this, we ask: \textit{which specific blocks establish the answer-relevant visual effect in text states, how much does each contribute, and how does this transfer relate to precision sensitivity and the necessity of subsequent visual access?}

To achieve this unified accounting, this paper proposes \emph{reciprocal nested intervention} (see Figure~\ref{fig:intro-transfer} 
for an illustration of our setup and notation). For two matched images under the same question, we run the forward pass of one image 
up to block $k$. We then replace its visual-token representations with those of the other image while retaining its text-token 
representations, and read out the final prediction. Averaging the effects across both directions yields the \emph{cumulative transfer} 
$C_k$: the portion of the total visual effect that has entered the text states by block $k$. By calculating adjacent differences 
$C_k-C_{k-1}$, we assign block-wise contributions that sum perfectly to the total visual effect. We term this layer-wise accounting 
the \emph{transfer ledger}. As illustrated in Figure~\ref{fig:intro-transfer}, this ledger yields two critical quantities for model 
compression: \emph{writer share} (identifying specific blocks whose visual-row precision should be protected) and \emph{causal closure} 
(predicting the earliest boundary where visual tokens can be safely evicted under the prompt-end readout).

Empirically, applying this approach to Qwen3.5-0.8B and -4B \citep{qwen35} and to LLaVA-1.5-7B \citep{liu2024llava15}, whose language 
backbone comes from the Llama family~\citep{touvron2023llama2}, reveals that transfer is highly concentrated. The two largest writer blocks alone carry $83\%$ and $71\%$ 
of the total visual effect in the two Qwen3.5 models, and the three largest carry $72\%$ in LLaVA-1.5-7B. This rapid accumulation establishes 
early causal closures at the inputs of blocks 12 and 16 (denoted $K12$ and $K16$) in the Qwen3.5 models, where visual tokens can be safely dropped 
while still retaining $95\%$ of the clean answer gap. This localized transfer also guides precision allocation. On Qwen3-VL-4B \citep{bai2025qwen3vl} 
and LLaVA-1.5-7B, the ledger-selected writers outperform blocks selected by attention-mass signals in the style of FastV and VTW. 
For the dominant writer $L7$ on Qwen3.5-0.8B, preserving only its $K/V$ projection inputs in BF16 recovers $94\%$ of the whole-block benefit.

However, causal closure is certified only for the first output token. During multi-token generation
on MMStar mathematics \citep{chen2024mmstar}, evicting visual tokens at $K16$ reduces Qwen3.5-4B
accuracy by $18$~pp, even though the ledger shows that the answer-relevant visual effect has already
entered the prompt's text states. Depth-truncation studies also find generation sensitive to image-token
removal \citep{ghosh2026process}, but truncation conflates stopping computation on visual tokens with
erasing them from memory. Separating the two, we find that generated tokens still read the visual rows
in late blocks but no longer need them to be updated. Simply freezing the rows at closure, which stops
their updates but keeps them readable, recovers $97\%$ of the eviction loss. Closure thus marks where
computation on the visual rows can stop, not where the rows can be discarded.

Our contributions are:
\begin{itemize}[leftmargin=1.4em,itemsep=1pt,topsep=1pt]
  \item We propose a quantitative accounting method to measure visual-to-text transfer, defining block-wise contributions and cumulative transfer against a unified answer-level effect.
  \item We find that visual-to-text transfer concentrates in a few writer blocks in both Qwen- and Llama-backbone VLMs, and that the ledger predicts compression sensitivity. Closure predicts the eviction boundary for short answers, and writer share identifies the blocks whose visual-row $K/V$ precision is worth protecting.
  \item We show that after closure, generated tokens still need to read the visual rows but no longer need them to be updated, so freezing the rows at closure instead of evicting them recovers most of the eviction loss on reasoning tasks.
\end{itemize}

%% file: sections/related_work.tex
\section{Related Work}
\input{sections/figure1}

\paragraph{Cross-modal information transfer in VLMs.}
Causal tracing places visual retrieval in early attention and feed-forward blocks and
attributes it to a few image tokens~\citep{basu2024storage}. Attention knockouts trace
image- and video-to-question transfer through shallow and middle
layers~\citep{zhang2025crossmodal,yin2025veil,kaduri2025whats,kim2026mapflow}, and later layers also read
object information directly from visual tokens~\citep{neo2025interpreting}. Closest
to ours, counterfactual patching shows that the balance between direct and text-mediated
routes depends on task, data, and prompt~\citep{salazar2026pathways}, and that image-patch
states stop affecting spatial answers after the middle layers~\citep{kang2026linear}. These
effects are measured separately at each location and do not sum to one answer-level total.
The ledger instead decomposes a common answer-level effect in a fixed order, linking
cumulative transfer, blockwise contributions, and residual visual dependence.

\paragraph{Visual-token reduction and precision sensitivity.}
FastV, SparseVLM, and PyramidDrop prune visual tokens using attention or similarity signals
inside the decoder~\citep{chen2024fastv,zhang2025sparsevlm,xing2025pyramiddrop}. Closest to
the question of when visual rows are no longer needed, VTW withdraws all visual tokens at a
depth calibrated by output-distribution shifts~\citep{lin2025vtw}, DyVTE learns an
input-dependent exit~\citep{wu2025dyvte}, and Information Horizon measures the prediction
effect of deleting tokens at each depth~\citep{wang2026horizon}. These criteria locate an
endpoint, whereas the ledger also accounts for how the answer effect was assembled before
it. ShortV instead freezes visual-token updates in layers selected by a per-layer
contribution score~\citep{yuan2025shortv}. During generation, depth truncation harms
multi-token outputs more than single-token predictions~\citep{ghosh2026process}, and the
needed visual evidence shifts as decoding proceeds~\citep{kim2026dstp}. Truncation, however,
both stops updating the visual tokens and removes them from later attention. Separating the
two shows that generation needs read access to the visual rows but not their updates, so
freezing the rows at the ledger's closure recovers most of the eviction loss.
Whereas MBQ, Q-VLM, and QIG calibrate uniform-precision quantization by modality balance,
cross-layer dependence, or gradient sensitivity~\citep{li2025mbq,wang2024qvlm,xiang2026qig},
the ledger prioritizes which blocks keep their visual-side $K/V$ inputs in BF16.

%% file: sections/figure1.tex
\begin{figure*}[t]
\centering
\includegraphics{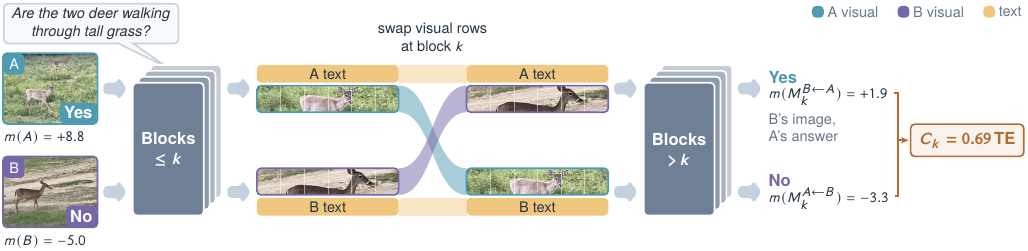}
\caption{Reciprocal nested intervention on a NaturalBench pair (Qwen3.5-4B), where $m$ is the
Yes--No log-likelihood margin at the end of the prompt. Values are shown for $k=11$, and $C_k$ is
computed from the two readouts by Eq.~\ref{eq:ck}.}
\label{fig:method-intervention}
\end{figure*}

%% file: sections/method_setup.tex
\section{Method}
\label{sec:method}

We measure visual-to-text transfer by replacing visual states with those from a matched 
reference image while retaining the text states, then measuring the answer contrast that remains
(Figure~\ref{fig:method-intervention}). We formalize this measurement as a transfer ledger with a 
closure criterion, validate the measurement, and derive testable predictions about visual-token
eviction (physical removal) and precision perturbations.

\subsection{Setup and the transfer estimand}
\label{sec:estimand}

\paragraph{Notation.}
An image $x$ is encoded into visual tokens, concatenated with the text tokens of a
question $q$, and passed through a decoder of $L$ blocks. Let $h_k$ be the hidden states
at the output of block $k$, $k \in \{-1,\dots,L-1\}$, with $h_{-1}$ the input embeddings.
Blocks are zero-indexed, so $L7$ denotes block 7 and $K12$ the input to block 12.
The rows are partitioned into \emph{visual rows} $V$, contiguous and fixed across depth,
and text rows. We write $h_{k,V}$ for the restriction to $V$, and the row class is the unit of
analysis throughout. Let $m(x,q)$ be a continuous answer score read out at the end of the prompt --- in the certified
instance below, the Yes/No log-likelihood margin on NaturalBench~\citep{li2024naturalbench}.

\paragraph{Reciprocal nested intervention.}
To instantiate this comparison, let $(A,B)$ be a semantically matched natural image pair sharing 
a question $q$, with directed total visual effect $\mathrm{TE} = s\,[\,m(A,q)-m(B,q)\,]$,
where $s\in\{\pm1\}$ aligns the semantic direction.
Their visual rows match in count and position.
The nested intervention $M_k^{B\leftarrow A}$ runs donor $A$ to $h_k^A$, replaces
\emph{only} $h_{k,V}^A$ with clean recipient states $h_{k,V}^B$, retains all text rows,
and runs the suffix to the readout. The reverse $M_k^{A\leftarrow B}$ is symmetric.
We define \emph{cumulative transfer} as the reciprocal, baseline-corrected contrast:
\begin{equation}
C_k \;=\; \frac{s}{2}\Big(
\big[\,m(M_k^{B\leftarrow A}) - m(B)\,\big]
-\big[\,m(M_k^{A\leftarrow B}) - m(A)\,\big]
\Big).
\label{eq:ck}
\end{equation}
$C_k$ is the part of the $A$/$B$ answer difference that has entered text states by
block $k$, and can therefore still be read out once the visual rows are reverted. Two
conventions anchor the ledger: the donor has not yet participated at $k=-1$, so $C_{-1}=0$;
and a restoration after the last block cannot affect the output, so $C_{L-1}=\mathrm{TE}$.

\paragraph{Writer contributions and exact accounting.}
Comparing adjacent intervention boundaries assigns the change in cumulative transfer to
block $k$: $\Delta_k = C_k - C_{k-1}$. Under this fixed order, interactions with earlier
states are attributed to the later block. Over a cohort of groups $g$ we report ratios of means,
\begin{equation}
r_k \;=\; \frac{\mathbb{E}_g[\Delta_{k,g}]}{\mathbb{E}_g[\mathrm{TE}_g]},
\qquad
F_k \;=\; \frac{\mathbb{E}_g[C_{k,g}]}{\mathbb{E}_g[\mathrm{TE}_g]},
\label{eq:shares}
\end{equation}
rather than averaging per-item ratios, which are unstable when the clean gap is near zero.
We call $r_k$ the \emph{writer share} of block $k$ and $F_k$ the \emph{completion} at $k$.
Both are signed and are not truncated to $[0,1]$, since under a fixed network order a block
may apply a reverse correction or settle an earlier overshoot. Telescoping
the ordered marginals between the two anchors yields an exact accounting identity,
\begin{equation}
\sum_{k=0}^{L-1}\Delta_k = \sum_{k=0}^{L-1}(C_k - C_{k-1}) = C_{L-1}-C_{-1} = \mathrm{TE},
\label{eq:telescope}
\end{equation}
so $\sum_{k} r_k = 1$, and the ledger accounts for the full visual effect under this ordered decomposition.

\paragraph{Causal closure.}
We next ask where dependence on visual rows becomes persistently negligible.
The cohort-level remaining visual dependence is $\rho_k = 1-F_k$.
We freeze a practical-equivalence band $\epsilon=0.05$ before
inspecting any layer scan, take a $90\%$ group-bootstrap interval for $\rho_k$, and define
the \emph{causal closure}
\begin{equation}
k^{*} \;=\; \min\Big\{\, k \;:\; \mathrm{CI}_{90}(\rho_j) \subseteq [-\epsilon,\,\epsilon]
\ \ \text{for all } j \ge k \,\Big\} ,
\label{eq:closure}
\end{equation}
with predicted eviction boundary $K^{*}=k^{*}{+}1$, where eviction at $K$ removes visual
tokens before block $K$. The criterion asserts \emph{equivalence} rather than absence
of significance: the entire interval must lie inside the frozen band. The suffix condition
$\forall j \ge k$ prevents a non-monotone curve from certifying a depth that later blocks
contradict, and $\epsilon$ is a fraction of the clean visual effect rather than of accuracy.
The ledger and closure are conditional on the model, stimulus distribution, and readout protocol.
Section~\ref{sec:limits} examines what this implies for generation.

\paragraph{Design choices.}
We use matched clean states rather than zero substitution, which can conflate
information removal with the model's response to an anomalous intervention. Across
Chameleon-7B~\citep{chameleon2024}, Qwen2.5-VL-3B~\citep{bai2025qwen25vl},
LLaVA-OneVision-0.5B~\citep{li2024llavaonevision}, and Qwen3.5-0.8B~\citep{qwen35},
zeroing the visual rows once and letting later blocks recompute them gives up to
$35$~pp lower accuracy than re-zeroing them after every block, and this gap is smaller
for every nonzero replacement we tried (Supp.~\ref{app:zero}). The matched donor also
gives the intervention a semantic direction and a corresponding total effect
$\mathrm{TE}$.

%% file: sections/method_validity.tex
\subsection{Instrument validity}
\label{sec:validity}

\begin{figure*}[t]
\centering
\includegraphics[width=\textwidth]{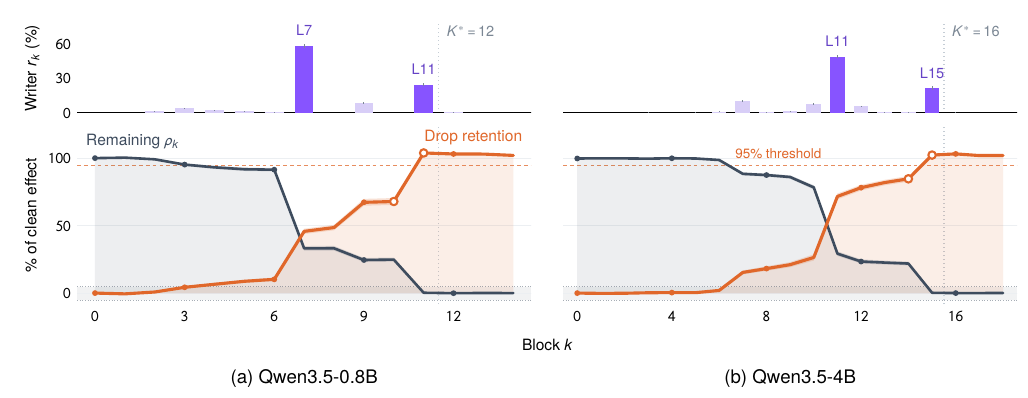}
\caption{Transfer ledgers and physical-drop validation on NaturalBench. A drop at the input of
block $K$ is plotted at $k=K-1$. Open circles mark $K^*-1$ and $K^*$, and the dotted line
marks $K^*$. The strip around zero is the equivalence band $\pm\epsilon$ for $\rho_k$. Intervals are
$90\%$ for $r_k$ and $\rho_k$ and $95\%$ for drop retention. Accuracy changes versus clean
(pp) at the circled boundaries are $-0.98$ ($K11$) and $+0.59$ ($K12$) for 0.8B, and $-3.13$
($K15$) and $+0.39$ ($K16$) for 4B.}
\label{fig:ledger}
\end{figure*}

The accounting identity holds by construction, so it cannot by itself show that
$\Delta_k$ measures transfer. We therefore check (a) that the implementation carries out
the intended intervention exactly, (b) that an independent intervention finds the same
writers, (c) that the predicted closure holds when visual tokens are physically
removed, and (d) that these results do not require matched donors.

\paragraph{(a) Implementation fidelity.}
We audit eight quantities at every group and boundary on Qwen3.5-0.8B and Qwen3.5-4B, with
zero tolerance: four algebraic endpoint checks and four implementation checks covering suffix 
equivalence, batch-path drift, visual-row restoration, and final recipient visual-state
agreement. All eight discrepancies
are exactly zero on both models (Supp.~\ref{app:audit} and~\ref{app:determinism}).

\paragraph{(b) Convergent validity.}
We next test whether $\Delta_k$ identifies the same writer locations as an independent intervention.
The \emph{edge-pulse} intervention inserts donor visual rows for one
block and restores recipient rows at its output. Unlike the donor-prefix marginal
$\Delta_k$, it isolates a single-block pulse, so we compare layer schedules rather
than absolute magnitudes. Layerwise means correlate with $\Delta_k$ at Pearson
$0.999$ ($0.8$B) and $0.991$ ($4$B), with exact recovery of the pre-registered
top-$k$ writers on both models (Supp.~\ref{app:edge}).

\paragraph{(c) Predictive validity under visual-token eviction.}
The ledger predicts closure from state replacement alone, without deleting visual tokens.
We test this prediction by eviction, measuring clean-gap retention as the mean
image-conditioned answer gap after eviction divided by its clean reference. On the same
NaturalBench cohort, the predicted $K12$ ($0.8$B) and $K16$ ($4$B) match the earliest
boundaries certified to retain at least $95\%$ of the clean gap, and all deeper tested boundaries
also pass (Figure~\ref{fig:ledger}). Accuracy changes there are $+0.59$ and $+0.39$~pp,
respectively. Full scans and certification criteria are in Supp.~\ref{app:kscan}.
Cross-cohort transfer is tested in Section~\ref{sec:signals}. Writer shares are
similarly concentrated on LLaVA-1.5-7B, where the three largest writers (L11, L8, L10)
carry $71.9\%$ of the total visual effect (Supp.~\ref{app:llava15}).

\paragraph{(d) Donor matching.}
Matched image pairs exist in only a few benchmarks, so we ask whether the ledger needs them. On
Qwen3.5-4B we replaced each matched donor with an image from another item, kept only if it
changes the answer and has the same visual-token layout, at both the original and native
resolution ($128$--$391$ visual tokens). In both settings $L11$ carries $47\%$ of the effect
($49\%$ with matched donors), the writer shares correlate with the matched ledger at Pearson
$0.98$, and closure stays at $K16$. At native resolution, eviction at $K16$ retains $100.8\%$ of
the clean gap, versus $88.0\%$ at $K15$. Only the ranking below $L11$ changes
(Supp.~\ref{app:donor}). The ledger therefore does not need matched pairs, only donors that
change the answer and share the recipient's token layout.

%% file: sections/method_predictions.tex
\subsection{From the ledger to compression predictions}
\label{sec:predictions}

We derive three predictions about compression from the transfer ledger before running the experiments in
Section~\ref{sec:experiments}. Throughout, quantization is simulated and applies only to visual-row activations.

\begin{itemize}[leftmargin=1.9em,itemsep=2pt,topsep=3pt]
  \item[\textbf{P1}] \emph{Eviction at the closure.}
Under the same readout, eviction at $K^*$ should preserve the clean visual effect,
whereas earlier eviction should incur loss. Section~\ref{sec:validity} tests this
prediction under the likelihood readout; Section~\ref{sec:signals} evaluates it on
generative benchmarks.
  \item[\textbf{P2}] \emph{Protection ordered by writer share.}
If $\Delta_k$ captures the transfer performed by block $k$, quantizing a large writer should 
be more damaging than quantizing a small one, so writer share should rank protection targets.
We test this ranking against attention-selected blocks and against non-writer blocks matched
for quantization-error energy.
  \item[\textbf{P3}] \emph{The $K/V$ port.}
Text queries read visual information through keys and values. With quantization confined
to visual rows, protecting the $K/V$ projection inputs should therefore recover most of
the whole-block protection benefit, whereas protecting $Q$ or $O$ should recover little.
\end{itemize}

Together, P1--P3 yield a simple phase rule: use
reduced precision before closure, prioritize high-share writers' $K/V$ inputs for precision
protection, and stop computing on visual tokens after closure. Whether the visual tokens can
also be evicted depends on the readout. An answer read at the end of the prompt no longer
needs them, but Section~\ref{sec:limits} shows that generated tokens still read them.

%% file: sections/experiments_setup.tex
\section{Experiments}
\label{sec:experiments}

\input{sections/table_compression}

\paragraph{Setup for Section~\ref{sec:signals}.}
We use two 4B decoders, Qwen3-VL-4B-Instruct~\citep{bai2025qwen3vl}, with
36 full-attention blocks, and Qwen3.5-4B~\citep{qwen35}, with 32 blocks that interleave
full and gated linear attention~\citep{yang2025gateddeltanet}, together with LLaVA-1.5-7B~\citep{liu2024llava15}, whose
32 full-attention blocks come from a Llama-family language model~\citep{touvron2023llama2}. We evaluate on
EmbSpatial-Bench~\citep{du2024embspatial}, POPE~\citep{li2023pope}, and
CV-Bench~\citep{tong2024cambrian}, reporting accuracy (F1 for POPE). Later subsections state
their own models and tasks. Each ledger comes from a matched-pair cohort as in
Section~\ref{sec:method} (the same NaturalBench cohort across all three models), and no benchmark result is used to choose
the boundary or the writers. Ledger construction caches boundary states and replays only each
suffix, requiring approximately $L{+}1$ forward-equivalents per item. The
NaturalBench runs take 24 and 22 minutes for Qwen3-VL-4B and Qwen3.5-4B, respectively,
on a single A100.

We evict visual tokens at each model's predicted boundary $K^*$ and protect
its largest-share writers (Table~\ref{tab:compression}): two blocks on the Qwen models and
three on LLaVA-1.5-7B, whose transfer is spread over more blocks. Attention-based selection
protects the same number of full-attention blocks with the largest visual attention mass, read
at the final prompt position as in FastV or averaged over text rows as in VTW
(Supp.~\ref{app:selectors}). A4, A3, and A2 quantize visual-row
activations to 4, 3, and 2 bits (asymmetric, per row) from block 3 through closure, and
\emph{protected} conditions keep the key and value projection inputs of the selected blocks
in BF16.

Table~\ref{tab:compression} scores the complete test sets. All runs use
lmms-eval~\citep{zhang2025lmmseval} on NVIDIA A100 GPUs with greedy decoding and identical
prompts within each model--benchmark pair. Decoding is deterministic, so each condition is
evaluated once and quoted differences are point estimates.

%% file: sections/table_compression.tex
\begin{table*}[t]
\centering
\small
\setlength{\tabcolsep}{10.3pt}
\caption{Ledger-guided compression on three decoders. Except in the first row, visual tokens
are evicted at the ledger's closure $K^*$. Attention rows report the better of the FastV and VTW
readings in each cell. $^\dagger$On Qwen3.5-4B both readings select the ledger blocks, so this row
protects the next two full-attention blocks by writer share. Bold marks the best protected result
per column and bit width.}
\label{tab:compression}
\begin{tabular}{ll ccc ccc ccc}
\toprule
& & \multicolumn{3}{c}{Qwen3-VL-4B ($K^*{=}20$)} & \multicolumn{3}{c}{Qwen3.5-4B ($K^*{=}16$)} & \multicolumn{3}{c}{LLaVA-1.5-7B ($K^*{=}14$)} \\
\cmidrule(lr){3-5}\cmidrule(lr){6-8}\cmidrule(lr){9-11}
\multicolumn{2}{l}{Ledger blocks}    & \multicolumn{3}{c}{13, 15} & \multicolumn{3}{c}{11, 15} & \multicolumn{3}{c}{8, 10, 11} \\
\multicolumn{2}{l}{Attention blocks} & \multicolumn{3}{c}{8, 15 (FastV) / 17, 18 (VTW)} & \multicolumn{3}{c}{3, 7$^\dagger$} & \multicolumn{3}{c}{4, 8, 13} \\
\midrule
Bits & Protected & EmbSp. & POPE & CV-B. & EmbSp. & POPE & CV-B. & EmbSp. & POPE & CV-B. \\
\midrule
\multicolumn{2}{l}{BF16, no eviction} & 76.90 & 89.28 & 85.25 & 75.33 & 89.01 & 85.37 & 45.25 & 81.92 & 56.89 \\
\multicolumn{2}{l}{Eviction at $K^*$} & 77.14 & 88.01 & 85.10 & 75.38 & 88.15 & 85.67 & 45.11 & 80.36 & 56.84 \\
\midrule
\multirow{3}{*}{A4} & None      & 73.13 & 79.95 & 81.46 & 69.09 & 81.58 & 78.96 & 43.93 & 78.07 & 56.96 \\
                    & Attention & 75.11 & 83.68 & 83.02 & 70.66 & 85.21 & 80.84 & 44.09 & 78.05 & 56.80 \\
\rowcolor{gray!12}  & Ledger    & \textbf{75.59} & \textbf{83.91} & \textbf{83.57} & \textbf{72.42} & \textbf{85.54} & \textbf{81.45} & \textbf{44.40} & \textbf{78.30} & \textbf{56.82} \\
\midrule
\multirow{3}{*}{A3} & None      & 57.31 & 30.40 & 53.94 & 46.46 & 38.33 & 52.23 & 41.90 & 48.02 & 55.34 \\
                    & Attention & 64.84 & 48.83 & 63.12 & 56.18 & 68.43 & 72.55 & 42.09 & 52.80 & 56.02 \\
\rowcolor{gray!12}  & Ledger    & \textbf{65.58} & \textbf{53.19} & \textbf{66.41} & \textbf{67.42} & \textbf{78.21} & \textbf{76.15} & \textbf{43.30} & \textbf{64.90} & \textbf{56.51} \\
\midrule
\multirow{3}{*}{A2} & None      & 42.47 & 0.00 & 42.00 & 39.64 & 0.66 & 39.76 & 36.40 & 0.00 & 45.54 \\
                    & Attention & 58.43 & 27.92 & 50.72 & 45.08 & 10.01 & 47.35 & 35.30 & 0.13 & 46.11 \\
\rowcolor{gray!12}  & Ledger    & \textbf{62.69} & \textbf{38.66} & \textbf{56.03} & \textbf{66.21} & \textbf{77.11} & \textbf{73.99} & \textbf{39.01} & \textbf{21.55} & \textbf{51.44} \\
\bottomrule
\end{tabular}
\end{table*}

%% file: sections/experiments_compression_signal.tex
\subsection{Causal transfer predicts compression-sensitive depths}
\label{sec:signals}

\begin{figure}[t]
\centering
\includegraphics[width=\columnwidth]{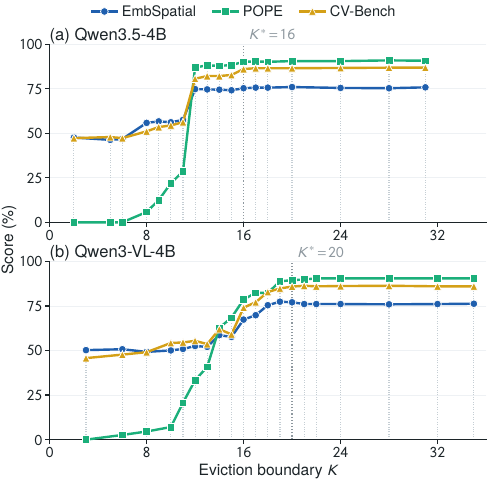}
\caption{Eviction-boundary scans on generative benchmarks (POPE scored by F1). The dotted line
marks the ledger's closure $K^*$.}
\label{fig:boundary-scan}
\end{figure}

\paragraph{Eviction boundary.}
To test the likelihood-calibrated closure on generative benchmarks, we sweep
visual-token eviction at block input $K$ with all other settings fixed.
Both checkpoints show stepwise recovery aligned with the ledger writers
(Figure~\ref{fig:boundary-scan}). On Qwen3.5, the boundary that first admits the dominant
writer $L11$ lifts all three benchmarks; on Qwen3-VL, the largest gains occur when blocks
13 and 15 first retain visual access. The ledger-predicted boundaries are the first
where all benchmarks come within $1.1$ points of their score at the deepest boundary
(Supp.~\ref{app:vl4b}) and remain there at
deeper boundaries. EmbSpatial on Qwen3.5 recovers earlier, so the boundary is conservative
for it. At these boundaries, eviction alone costs at most $1.6$ points on any benchmark
(Table~\ref{tab:compression}). Neither ledger saw these items or
generative scoring, so the agreement is out-of-sample support for the likelihood-based
certificate.

\paragraph{Writer selection.}
To test \hyperref[sec:predictions]{P2}, we fix the boundary, bit width, and protection
budget, varying only the protected blocks. Quantization costs grow steeply as precision
falls, and at A2 POPE F1 drops to near zero on all three models. The ledger-selected blocks lead
in every cell of Table~\ref{tab:compression}. Margins stay within $2$ points under A4 and
widen as precision falls, reaching tens of points on POPE at A2. On Qwen3.5-4B, both attention
readings select the ledger pair. Protecting the next two full-attention blocks instead leaves
A2 POPE F1 at $10.01$, versus $77.11$ for the ledger pair. Attention recovers the writers on one of the three checkpoints, whereas
the ledger ranks the blocks worth protecting on all three.

%% file: sections/exp_ablations.tex
\subsection{What explains writer protection}
\label{sec:ablations}

All ablations use Qwen3.5-0.8B, whose predicted eviction boundary at $K12$ was validated in
Section~\ref{sec:validity}. Visual-row activations are quantized over $L3$--$L11$,
and protected inputs remain in BF16. We first re-run the ledger on the NaturalBench
cohort to measure how quantization changes transfer. Protection is then evaluated on
$2{,}944$ A-OKVQA~\citep{schwenk2022aokvqa} questions (four-choice likelihood, eviction at $K12$;
Table~\ref{tab:ablation}), testing whether its benefit tracks writer identity, which
input port carries it, and whether weak precision sensitivity implies dispensability.

\paragraph{Quantization degrades transfer before it degrades accuracy.}
Under A4 quantization, the total visual effect has already fallen by a fifth, while accuracy shows 
no detectable change, so task accuracy can conceal mechanistic degradation. At 
A2 only an eighth of the effect survives and the schedule is rerouted: its layerwise 
correlation with BF16 falls to $0.18$, and the top writers move from $L7/L11/L9$
to $L9/L4/L5$. Dose curves for accuracy and writer marginals are in Supp.~\ref{app:dose}.

\begin{table}[t]
\centering\small
\setlength{\tabcolsep}{5pt}
\caption{Accuracy recovered by keeping only the listed inputs in BF16 (Qwen3.5-0.8B,
A-OKVQA, $n{=}2{,}944$). Visual-row activations over $L3$--$L11$ are quantized to A2, and
visual tokens are evicted at $K12$. Without protection, this setting trails BF16 by
$17.29$~pp. Recovery is measured against the unprotected setting and is also given as a
fraction of this gap. Energy is the protected inputs' share of the window's
quantization-error energy. $L4{+}L5$ and $L8{+}L9$ are Gated DeltaNet (GDN) pairs on either
side of $L7$, matched to each other in error energy. $^\dagger$Not significant under
family-wise max-T correction (Supp.~\ref{app:ablation}).}
\label{tab:ablation}
\begin{tabular}{lrrr}
\toprule
Protected inputs & Energy & Recov.\ (pp) & \% of gap \\
\midrule
$L7$ (writer)            & 17.6\% & $+7.64$          & 44\% \\
$L3$ (non-writer)        & 21.0\% & $+2.68$          & 16\% \\
$L4{+}L5$ (early GDN)    & 13.2\% & $+1.56$          & 9\%  \\
$L8{+}L9$ (late GDN)     & 14.0\% & $+0.44^\dagger$  & 3\%  \\
\midrule
$L7$ attention           & 10.3\% & $+7.24$          & 42\% \\
$L7$ MLP                 & 7.3\%  & $+0.41^\dagger$  & 2\%  \\
\rowcolor{gray!12}
$\boldsymbol{L7}$ $\boldsymbol{K/V}$ \textbf{only}         & 6.8\%  & $\boldsymbol{+7.17}$  & 41\% \\
$L7$ $Q$ only            & 3.4\%  & $-0.07^\dagger$  & 0\%  \\
$L11$ $K/V$ only         & 8.7\%  & $+3.64$          & 21\% \\
\bottomrule
\end{tabular}
\end{table}

\paragraph{Writer identity survives matched controls (P2).}
Table~\ref{tab:ablation} fixes the quantization 
setting and varies only which inputs stay in BF16. Protecting writer \(L7\) recovers \(4.96\)~pp more 
than \(L3\), despite the same module type and greater quantization-error energy at \(L3\). For depth, we 
interpolate the recovery trend between two error-energy-matched Gated DeltaNet (GDN) pairs bracketing 
\(L7\) (\(L4{+}L5\) and \(L8{+}L9\)). Writer \(L7\) exceeds this trend by \(6.78\)~pp. Both contrasts are max-T 
significant at \(p<10^{-5}\) (Supp.~\ref{app:ablation}), so the benefit tracks writer identity 
beyond module type, error energy, and depth. Between the two writers, the larger is also the
more valuable target: \(L7\)'s \(K/V\) inputs recover twice as much as \(L11\)'s.

\paragraph{The benefit is carried by the $K/V$ port (P3).}
Protecting only $L7$'s $K/V$ inputs recovers $94\%$ of the whole-block benefit,
although these activations account for just $6.8\%$ of the window's
quantization-error energy. The MLP and $Q$ together hold most of the block's
error energy, yet neither yields significant recovery after max-T correction
(Table~\ref{tab:ablation}), nor does $O$ (Supp.~\ref{app:ablation}).
Error energy alone does not explain the protection pattern across
blocks or within $L7$. Keeping the writers' attention weights in BF16 instead shows no
reliable writer-specific benefit (Supp.~\ref{app:negative}), so the rule protects
activations only.

\paragraph{Presence matters more than precision.}
The small benefit from protecting MLP precision does not mean the visual MLP is dispensable:
disabling it entirely costs $10.94$~pp where running it at A4 costs $2.34$~pp in the same
condition. Before closure, the prescription is therefore reduced precision, not skipped
computation.

%% file: sections/exp_limits.tex
\subsection{Closure ends computation, not access}
\label{sec:limits}

\begin{table}[b]
\centering\small
\setlength{\tabcolsep}{2.7pt}
\caption{Separating computation from access during generation (MMStar accuracy, \%, $100$
items per task). From closure on ($K16$ for Qwen3.5-4B, $K20$ for Qwen3-VL-4B), \emph{Comp.}
marks whether the visual rows keep being updated and \emph{Read} whether generated tokens can read
them. Evict and Freeze intervene from prefill on, whereas Block reads keeps prefill and the first
output token clean.}
\label{tab:access}
\begin{tabular}{l cc ccc ccc}
\toprule
& & & \multicolumn{3}{c}{Qwen3.5-4B} & \multicolumn{3}{c}{Qwen3-VL-4B} \\
\cmidrule(lr){4-6}\cmidrule(lr){7-9}
Condition & Comp. & Read & Math & Log. & Sci. & Math & Log. & Sci. \\
\midrule
Full        & \cmark & \cmark & 76 & 70 & 47 & 72 & 73 & 38 \\
Evict       & \xmark & \xmark & 57 & 56 & 49 & 50 & 58 & 37 \\
Block reads & \cmark & \xmark & 57 & 56 & 49 & 55 & 64 & 39 \\
\rowcolor{gray!12}
\textbf{Freeze} & \xmark & \cmark & \textbf{75} & \textbf{68} & \textbf{49} & \textbf{69} & \textbf{72} & \textbf{37} \\
\bottomrule
\end{tabular}
\end{table}

\begin{figure}[t]
\centering
\includegraphics[width=0.96\columnwidth]{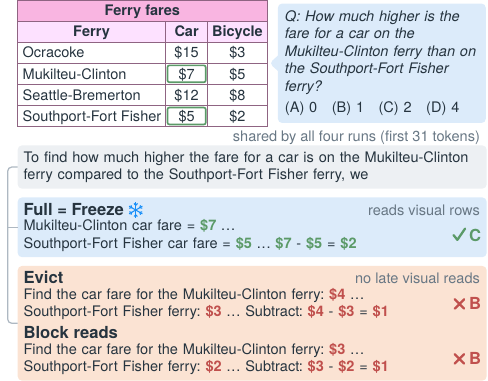}
\caption{A generation that needs late visual reads (MMStar, Qwen3-VL-4B). All four runs generate
the same first 31 tokens. With read access to the visual rows (Full and Freeze, whose outputs are
identical), the model looks up the car fares \$7 and \$5. Without late reads (Evict, Block reads),
it reports wrong fares. Block reads returns \$3 and \$2, which are in the table but in the bicycle
column. Outputs are excerpted, and the table is re-typeset for legibility from the
$349{\times}187$ input image.}
\label{fig:case}
\vspace{-1.0em}
\end{figure}

\paragraph{Eviction fails although transfer is complete.}
Under generative evaluation, eviction at $K16$ lowers Qwen3.5-4B's MMStar~\citep{chen2024mmstar}
accuracy by $18.0$~pp in mathematics and $13.6$~pp in logical reasoning, while the other four
capabilities show no detectable change (Supp.~\ref{app:readout}). Transfer is nonetheless
complete: on mathematics items only $1.3\%$ of the visual effect remains at $K16$, and
reasoning-heavy items complete their transfer earlier than perception items
(Supp.~\ref{app:taskmap}).
Holding prefill and the first output token clean, we keep all block computations intact and
block only generated tokens from reading late-layer visual keys and values. This alone lowers
mathematics accuracy by $19$~pp on Qwen3.5-4B and $17$~pp on Qwen3-VL-4B
(Table~\ref{tab:access}; Supp.~\ref{app:decode-visual}). Figure~\ref{fig:case} shows one such item, in
which the model reads the wrong cells of a table. Nearly complete prompt-side
transfer therefore does not remove the need for later visual access.

\paragraph{What generated tokens read.}
We then checked whether the facts that reasoning chains rely on had already entered the
prompt. After the clean prefill, we asked Qwen3-VL-4B for four such facts (a table entry, a
marked length, a count, and a volume), with the new question reading only the prompt's text
states. It answered none correctly, in three cases repeating its no-image answer, yet
answered all four with visual access or when the fact was written into the prompt
(Supp.~\ref{app:decode-visual}). The details that reasoning chains retrieve are therefore
not readable from the prompt's text states and must come from the visual rows themselves.

\paragraph{Freezing restores access without computation.}
We therefore keep the visual rows readable but stop updating them~\citep{yuan2025shortv} at closure: later blocks no
longer compute on these rows, yet every later query can still attend to them as keys and
values. Freezing recovers $97\%$ of the
eviction loss on Qwen3.5-4B and $87\%$ on Qwen3-VL-4B, and leaves science items, which
eviction does not hurt, unchanged. Across the four conditions, accuracy follows whether
generated tokens can read the visual rows, not whether the rows are still computed
(Table~\ref{tab:access}). After closure, frozen rows need only their $K/V$ projections, about
$5\%$ of a block's per-token linear cost, so freezing keeps most of eviction's computation
savings while giving up its KV-cache savings.

\begin{figure}[t]
\centering
\includegraphics[width=0.96\columnwidth]{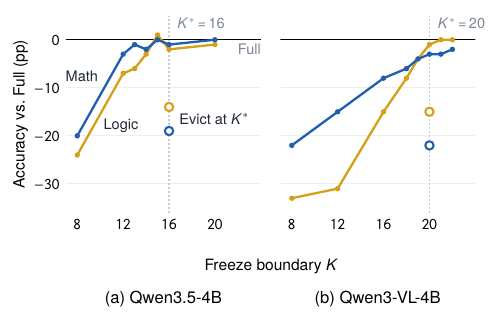}
\caption{Freeze-boundary sweep on MMStar (100 items per task). Freezing at $K$ stops updating the
visual rows from block $K$ on but keeps them readable. Hollow markers show eviction at $K^*$, and
the dotted line marks $K^*$.}
\label{fig:freeze}
\vspace{-1.0em}
\end{figure}

\paragraph{Moving the freeze boundary.}
While Table~\ref{tab:access} freezes the rows at closure, Figure~\ref{fig:freeze} sweeps this
boundary across depth.
Freezing at $K8$ lowers accuracy by $20$--$33$~pp, as much as eviction at closure or more, and
accuracy rises as the boundary moves deeper. Unlike eviction at $K$, freezing from $K$ leaves
block $K$'s reads unchanged, so the freeze boundary matching closure is $K^*{-}1$, namely $K15$ on
Qwen3.5-4B and $K19$ on Qwen3-VL-4B. There, freezing comes within $4$~pp of full accuracy, whereas
eviction at closure loses $14$--$22$~pp.

%% file: sections/conclusion.tex
\section{Conclusion}
\label{sec:conclusion}

In this work, we introduce an answer-level accounting framework for visual-to-text transfer,
turning layerwise interventions into an exact decomposition of how visual evidence
is established in text states. Across models and tasks, this accounting shows that
transfer is concentrated in a few writer blocks and links this organization to
compression sensitivity. It also reveals that after closure, the visual rows need no more
computation but must stay readable during generation. More broadly,
this view turns visual-token eviction from a depth-selection problem into a question 
of how visual evidence is established, preserved, and subsequently accessed.

%% file: sections/appendix.tex
\raggedbottom
\providecommand{\maincite}[1]{\citep{#1}}
\providecommand{\cisub}[1]{\raisebox{-0.2ex}{\scriptsize\color{black!55}#1}}
\makeatletter
\providecommand{\supp@floatsheld}[2]{\begingroup\let\@elt\relax
  \edef\@tempa{\@botlist\@deferlist\@dbldeferlist}%
  \ifx\@tempa\@empty\endgroup#2\else\endgroup#1\fi}
\providecommand{\suppfloatbarrier}{\par
  \supp@floatsheld{\newpage\supp@floatsheld{\clearpage}{}}{}}
\makeatother
\providecommand{\valci}[2]{\begin{tabular}[t]{@{}c@{}}#1\\[-2.5pt]{\scriptsize\color{black!55}#2}\end{tabular}}
\providecommand{\supptocline}[1]{\hyperref[#1]{\makebox[1.6em][l]{\ref*{#1}}\nameref*{#1}%
  \enspace\dotfill\enspace\pageref*{#1}}\par}

\begingroup
\long\def\texorpdfstring#1#2{#2}
\begin{center}
\begin{minipage}{0.8\linewidth}
\small
\textbf{Contents}\par\smallskip
\supptocline{app:determinism}
\supptocline{app:zero}
\supptocline{app:audit}
\supptocline{app:edge}
\supptocline{app:kscan}
\supptocline{app:donor}
\supptocline{app:vl4b}
\supptocline{app:dose}
\supptocline{app:ablation}
\supptocline{app:negative}
\supptocline{app:taskmap}
\supptocline{app:decode-visual}
\end{minipage}
\end{center}
\endgroup

\section{Determinism and statistical conventions}
\label{app:determinism}

Each item's paired conditions run in the same batch on the same device. Framework
determinism flags alone did not make margins reproducible, so we also pin the FLA
autotuner and require two full-grid runs per device to agree exactly.

\paragraph{Shared statistical protocol.}
\label{app:statistics}
Unless a section states otherwise, intervals are pointwise $95\%$ paired bootstrap
intervals from $10{,}000$ item- or group-level resamples that keep all conditions of an item
together, stratified by device in the protection and task-map experiments. Family-wise
$p$-values use two-sided max-T over $200{,}000$ sign-flip draws. Intervals containing zero are reported as no
detected difference rather than equivalence.

\section{Zero substitution}
\label{app:zero}

Zeroing the visual rows is not a neutral removal: letting later blocks recompute the zeroed
rows (release) gives lower accuracy than re-zeroing them after every block (clamp) on all four
models, and no nonzero replacement we tested shows a comparable gap.

\paragraph{Interventions.}
At the output of zero-indexed block $b$, $Z$ zeros the visual rows once (release),
$Z_{\mathrm{pers}}$ re-zeros them after every block $\ell\ge b$ (clamp), and $Z_{\mathrm{mask}}$
applies the same initial zeroing and then masks visual keys with $-\infty$ in every later
full-attention block, leaving GDN state pathways unmasked. With $a(\cdot)$ the accuracy in pp,
\[
T=a(Z_{\mathrm{mask}})-a(Z),\qquad P=a(Z_{\mathrm{mask}})-a(Z_{\mathrm{pers}}),\qquad
R=a(Z_{\mathrm{pers}})-a(Z),\quad T=P+R.
\]
We evaluate $256$ A-OKVQA items (indices $256$--$511$) with BF16 eager inference and
four-choice length-normalized likelihood.

\paragraph{Mask, clamp, and release.}
Table~\ref{tab:zero-contrasts} reports one block per model on Chameleon-7B~\maincite{chameleon2024},
Qwen2.5-VL-3B~\maincite{bai2025qwen25vl}, LLaVA-OneVision-0.5B~\maincite{li2024llavaonevision},
and Qwen3.5-0.8B, chosen to illustrate distinct patterns. The clamp advantage $R$ is
positive on all four and reaches $35$~pp on Qwen2.5-VL-3B, while $P$ can reinforce or
oppose it. A zeroing curve alone therefore cannot separate semantic dependence from the
model's response to the intervention.

\begin{table}[!htbp]
\centering\small
\caption{A-OKVQA accuracy contrasts ($n=256$ per model): mask-minus-release ($T$),
mask-minus-clamp ($P$), and clamp-minus-release ($R$). Entries are pp with pointwise
$95\%$ paired-bootstrap CIs in gray; intervention blocks are zero-indexed.}
\label{tab:zero-contrasts}
\begin{tabular}{l r@{\,}l r@{\,}l r@{\,}l}
\toprule
Model/block & \multicolumn{2}{c}{$T$ (pp)} & \multicolumn{2}{c}{$P$ (pp)} & \multicolumn{2}{c}{$R$ (pp)} \\
\midrule
Chameleon-7B $L15$        & $+22.27$ & \cisub{$[+14.84,+29.69]$} & $-1.95$ & \cisub{$[-6.25,+2.34]$}   & $+24.22$ & \cisub{$[+16.41,+31.64]$} \\
Qwen2.5-VL-3B $L19$       & $+28.12$ & \cisub{$[+21.09,+35.16]$} & $-7.03$ & \cisub{$[-12.89,-1.56]$}  & $+35.16$ & \cisub{$[+27.73,+42.58]$} \\
LLaVA-OneVision-0.5B $L9$ & $+12.89$ & \cisub{$[+7.81,+17.97]$}  & $+7.42$ & \cisub{$[+2.34,+12.50]$}  & $+5.47$  & \cisub{$[+0.78,+10.16]$}  \\
Qwen3.5-0.8B $L5$         & $+1.95$  & \cisub{$[-1.56,+5.47]$}   & $-2.73$ & \cisub{$[-5.47,-0.39]$}   & $+4.69$  & \cisub{$[+1.17,+8.20]$}   \\
\bottomrule
\end{tabular}
\end{table}

\paragraph{Replacement controls.}
For each operator $o$, one-shot replacement acts once after block $b$, and persistent
replacement re-applies the layer-specific target $\tilde{h}_\ell^o$ after every block $\ell\ge b$,
with response $S_o=a(\mathrm{persistent}(o))-a(\mathrm{one}(o))$, so that $S_{\mathrm{zero}}=R$.
\emph{Mean} broadcasts per-layer, per-channel visual-row means from $128$ calibration images,
\emph{Noise} draws Gaussian rows matched to those means and standard deviations,
\emph{Cross-50\%} patches a fixed random half of positions with same-position clean donor
rows, and \emph{Perm} applies a fixed row permutation to the recipient's clean trajectory.
We run these follow-up controls on three of the four models---Chameleon-7B ($n=256$),
Qwen2.5-VL-3B, and Qwen3.5-0.8B ($n=252$, excluding four items whose image grid has no
same-size donor)---spanning the full range of clamp responses ($+4.7$ to $+35.2$~pp), and omit
LLaVA-OneVision-0.5B. All twelve nonzero responses in Table~\ref{tab:replacement-controls}
are smaller than the zero response, with differences $S_o-S_{\mathrm{zero}}$ from $-4.37$ to
$-39.45$~pp and every interval below zero.

\begin{table}[!htbp]
\centering\small
\caption{Persistent-minus-one-shot accuracy for zero and nonzero replacements on A-OKVQA:
$n=256$ for Chameleon and $n=252$ for each Qwen model. Entries give $S_o$ in pp with
pointwise $95\%$ paired-bootstrap CIs shown below; positive values favor persistent replacement.}
\label{tab:replacement-controls}
\begin{tabular}{@{}lccccc@{}}
\toprule
Model/block & Zero & Mean & Noise & Cross-50\% & Perm \\
\midrule
Chameleon-7B $L15$  & \valci{$+24.22$}{$[+16.41,+31.64]$} & \valci{$-15.23$}{$[-23.44,-7.03]$} & \valci{$+0.78$}{$[-5.47,+7.42]$}  & \valci{$+0.39$}{$[-0.78,+1.95]$} & \valci{$-0.78$}{$[-2.34,+0.78]$} \\[3pt]
Qwen2.5-VL-3B $L19$ & \valci{$+36.11$}{$[+28.57,+43.65]$} & \valci{$-0.40$}{$[-3.17,+2.38]$}   & \valci{$+0.40$}{$[-1.98,+2.78]$}  & \valci{$0.00$}{$[-1.19,+1.19]$}  & \valci{$+0.40$}{$[0.00,+1.19]$}  \\[3pt]
Qwen3.5-0.8B $L5$   & \valci{$+4.76$}{$[+1.59,+8.33]$}    & \valci{$-1.59$}{$[-6.35,+3.17]$}   & \valci{$-3.97$}{$[-7.14,-1.19]$}  & \valci{$-0.40$}{$[-1.98,+1.19]$} & \valci{$+0.40$}{$[-1.59,+2.38]$} \\
\bottomrule
\end{tabular}
\end{table}

\section{Audit quantities}
\label{app:audit}

The eight audit discrepancies of Section~\ref{sec:validity} are exactly zero on
Qwen3.5-0.8B and Qwen3.5-4B. The grids contain $256$ groups with two questions each and
$25$ or $33$ boundaries ($12{,}800$ and $16{,}896$ rows). Maxima run over groups, questions,
reciprocal directions, and boundaries (indices suppressed). $\mathbf z$ denotes the
answer-scoring logits and $\mathbf u=\operatorname{Norm}(\mathbf h_{L-1})$ the final
normalized hidden states returned by the suffix.

\begin{equation*}
\begin{aligned}
E_{\mathrm{start}} &= \max_g |C_{-1,g}|, &
E_{\mathrm{end}} &= \max_g |C_{L-1,g}-\mathrm{TE}_g|, \\
E_{\mathrm{tel}} &= \max_g \big|\textstyle\sum_k\Delta_{k,g}-(C_{L-1,g}-C_{-1,g})\big|, &
E_{\mathrm{ledger}} &= \max_g \big|\mathrm{TE}_g-\textstyle\sum_k\Delta_{k,g}\big|, \\
E_{\mathrm{suffix}} &= \max_{g,k}\lVert \mathbf z_{g,k}^{\mathrm{clean\text{-}suffix}}-\mathbf z_g^{\mathrm{full}}\rVert_\infty, &
E_{\mathrm{dup}} &= \max_{g,k}\lVert \mathbf h_{g,k}^{\mathrm{dup}}-\mathbf h_{g,k}^{\mathrm{ctrl}}\rVert_\infty, \\
E_{\mathrm{restore}} &= \max_{g,k}\lVert \widetilde{\mathbf h}_{g,k,V}-\mathbf h_{g,k,V}^{\mathrm{recip}}\rVert_\infty, &
E_{\mathrm{traj,final}} &= \max_{g,k}\lVert \widetilde{\mathbf u}_{g,V}^{(k)}-\mathbf u_{g,V}^{\mathrm{recip}}\rVert_\infty.
\end{aligned}
\end{equation*}

The first four quantities check the accounting algebra, and the other four check that
suffix replay reproduces the full forward pass ($E_{\mathrm{suffix}}$), that duplicated
donor-prefix states match the control ($E_{\mathrm{dup}}$), that restored visual rows equal
the recipient's ($E_{\mathrm{restore}}$), and that the recipient's final visual states are
recovered ($E_{\mathrm{traj,final}}$). The restoration step leaves text rows untouched and
recovers the recipient's clean visual trajectory---because text follows vision under causal
ordering, altered text states cannot feed back into visual rows---with suffix states
recomputed rather than reusing donor caches.

\section{Edge-pulse convergence}
\label{app:edge}

\paragraph{Protocol.}
For every block $k$, we run the clean recipient $B$ to the input of block $k$, replace only
its visual rows with those of donor $A$, and run block $k$ so $A$'s visual evidence can
reach the text states. We then restore $B$'s visual rows at the block output while keeping
the updated text states, run the clean suffix, and form the reciprocal effect as in
Eq.~\ref{eq:ck}. Both models use the $256$-group NaturalBench cohort ($512$ reciprocal
questions), BF16 eager inference, thinking disabled, and $121$ visual tokens.

\begin{table}[!htbp]
\centering\small
\caption{Edge-pulse and nested schedules on NaturalBench ($256$ groups per model).
Pearson correlation compares layerwise means; the final column reports top-three overlap.}
\label{tab:edge-convergence}
\begin{tabular}{llrlll}
\toprule
Model & Blocks & Pearson & Edge top-3 & Nested top-3 & Result \\
\midrule
Qwen3.5-0.8B & 24 & $0.999$ & $L7/L11/L9$ & $L7/L11/L9$ & 3/3, exact \\
Qwen3.5-4B & 32 & $0.991$ & $L11/L15/L7$ & $L11/L15/L7$ & 3/3, exact \\
\bottomrule
\end{tabular}
\end{table}

Our pre-registered recovery criterion required the top two edge writers in the nested top-3
($L7, L11$ on $0.8$B; $L11, L15$ on $4$B), the third ($L9$; $L7$) in the top-5, and at least
two recovered; both models achieve an exact 3/3 match (Table~\ref{tab:edge-convergence}).

\paragraph{Layerwise comparison.}
Figure~\ref{fig:edge-scan} normalizes both schedules by the nested run's mean clean total
effect. Nested shares exceed edge shares at early writers and nearly coincide at the final
writer, as expected when the ordered marginal assigns donor-prefix interactions to the
current block.

\input{sections/edge_scan_figure}

\section{Eviction validation}
\label{app:kscan}

Across six decoders, the ledger-predicted boundary $K^*$ is never earlier than the earliest
certified-safe boundary $K_{\mathrm{safe}}$: the two coincide on four models, and $K^*$ is one
block conservative on the other two (Table~\ref{tab:ledger-summary}). This section gives the
full eviction scans behind Section~\ref{sec:validity} and details the four further decoders.

\paragraph{Shared criterion.}
Visual tokens are evicted at the input of block $K$, keeping the original (M)RoPE position
ids of the surviving rows. A boundary is certified safe if the lower bound of the retained clean gap is
at least $95\%$, materially unsafe if the upper bound is below $95\%$, and indeterminate
otherwise. Each run is normalized by its own clean reference, including in
Figure~\ref{fig:ledger}. Accuracy changes are reported for reference and do not define
safety. $K_{\mathrm{safe}}$ denotes the earliest boundary whose entire tested suffix is
certified safe.

\begin{table}[!htbp]
\centering\small
\caption{Ledger and eviction summary across decoders. Top-3 share is the summed writer share
of the three largest writers; $K^*$ is the ledger-predicted eviction boundary and
$K_{\mathrm{safe}}$ the earliest suffix-safe boundary. The two differ only on Qwen3-VL-2B
and LLaVA-1.5-7B, where $K^*$ is one block later.}
\label{tab:ledger-summary}
\begin{tabular}{lllrrr}
\toprule
Model & LM backbone & Top-3 writers & Top-3 share & $K^*$ & $K_{\mathrm{safe}}$ \\
\midrule
Qwen3.5-0.8B  & Qwen3.5 (hybrid) & $L7$, $L11$, $L9$   & $92\%$ & $12$ & $12$ \\
Qwen3.5-4B    & Qwen3.5 (hybrid) & $L11$, $L15$, $L7$  & $81\%$ & $16$ & $16$ \\
Qwen3-VL-2B   & Qwen3            & $L11$, $L12$, $L13$ & $91\%$ & $15$ & $14$ \\
Qwen3-VL-4B   & Qwen3            & $L15$, $L13$, $L19$ & $67\%$ & $20$ & $20$ \\
LLaVA-OV-0.5B & Qwen2            & $L9$, $L12$, $L11$  & $78\%$ & $13$ & $13$ \\
LLaVA-1.5-7B  & Vicuna (Llama-2) & $L11$, $L8$, $L10$  & $72\%$ & $14$ & $13$ \\
\bottomrule
\end{tabular}
\end{table}

\subsection{Qwen3.5 on NaturalBench}
\label{app:drop-qwen35}

Figure~\ref{fig:drop-scan} shows accuracy changes at all scanned boundaries. Under the retention criterion, every $K<K^*$ is materially unsafe
and every $K\geq K^*$ certified safe, with $K^*=12$ over $K1$--$K23$ on $0.8$B and $K^*=16$
over $K1$--$K31$ on $4$B.
\input{sections/drop_scan_extrema}

\input{sections/drop_scan_figure}

\subsection{Qwen3-VL and LLaVA-OneVision}
\label{app:vl-drop}

The Qwen3-VL-2B/4B-Instruct runs use the same $256$-group NaturalBench cohort, prompt, and
readout as the Qwen3.5 ledgers, and scan every block input from $K3$ (after DeepStack
injection) to the last block. On Qwen3-VL-4B (intervals in Figure~\ref{fig:ledger-vl}a), every
$K\leq18$ incurs material loss and $K19$ is indeterminate at $96.1\%$ $[94.8, 97.3]$
($-1.97$~pp). Certification starts at the predicted $K20$ ($101.0\%$, $+0.12$~pp), and
admitting $L13$, $L15$, and $L18$ gives the three largest single-block gains.

Two additional decoders summarized in Table~\ref{tab:ledger-summary}---Qwen3-VL-2B and
LLaVA-OneVision-0.5B~\maincite{li2024llavaonevision}---follow the same pattern. On
Qwen3-VL-2B, retention first passes at $K14$ ($99.3\%$, lower bound $98.0\%$) while accuracy
is still $-3.59$~pp, and accuracy recovers at the predicted $K15$. On LLaVA-OneVision-0.5B,
$L9$ alone carries $51.2\%$ of the effect, and the predicted $K13$ is the earliest certified
boundary ($99.0\%$, lower bound $98.3\%$; $-0.39$~pp).

\subsection{LLaVA-1.5-7B}
\label{app:llava15}

\begin{figure}[t]
\centering
\includegraphics[width=0.95\linewidth]{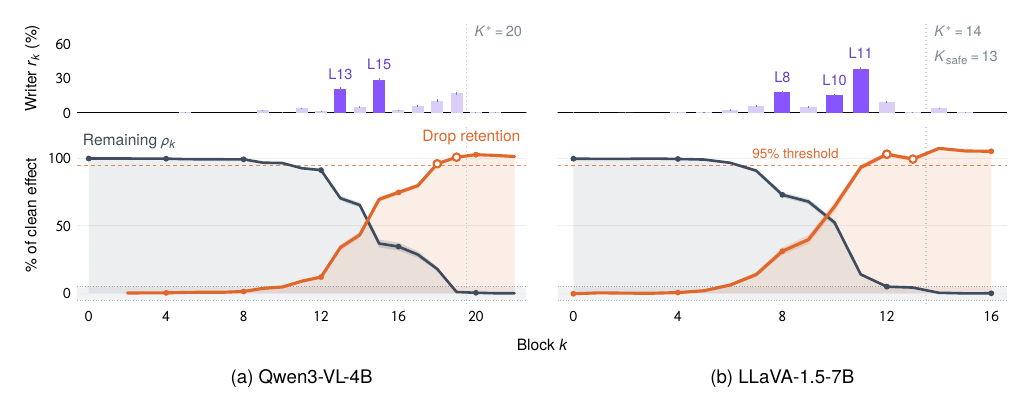}
\caption{Transfer ledgers and eviction validation on Qwen3-VL-4B and LLaVA-1.5-7B. Writer
shares and remaining dependence have $90\%$ intervals, and drop retention has $95\%$
intervals. A drop at input $K$ is plotted at $k=K-1$, and dotted vertical lines mark $K^*$.
The band around zero is the $\pm5\%$ closure band, and the dashed line is the $95\%$
retention threshold. Panels show three blocks after closure; every deeper tested boundary
also passes.}
\label{fig:ledger-vl}
\end{figure}

LLaVA-1.5-7B~\maincite{liu2024llava15} pairs a CLIP vision encoder with a $32$-block
full-attention Vicuna (Llama-2) decoder. On the same $256$-group NaturalBench cohort, inputs
are resized to $336\times336$ ($576$ visual rows) and prompted with the Vicuna-v1 template
(``Yes, No'' options), scored by the Yes/No single-token log-likelihood margin under BF16
eager inference; all eight audits of Supp.~\ref{app:audit} are exactly zero.

Transfer concentrates in $L6$--$L14$ ($98.8\%$ of the total effect; top-3 writers $L11$,
$L8$, and $L10$ carry $71.9\%$, Table~\ref{tab:ledger-summary} and
Figure~\ref{fig:ledger-vl}b). Remaining dependence already falls to $4.9\%$ after $L12$, but
its $90\%$ upper bound ($5.7\%$) exceeds the closure band, certifying closure one block later
at $k^*=13$ ($K^*=14$). Under eviction, every $K\leq11$ incurs material loss, $K12$ is
indeterminate ($93.4\%$ $[91.1, 95.6]$), and $K13$--$K31$ are certified safe; at the
predicted $K14$, retention is $99.6\%$ with $-0.10$~pp accuracy change, making $K^*$ one
block conservative.

\section{Ledgers with unpaired donors}
\label{app:donor}

\paragraph{Donor selection.}
On Qwen3.5-4B, each recipient is the first image of one of the $256$ NaturalBench groups
with its two questions, and donors are drawn from all $1{,}450$ NaturalBench Yes/No groups
($2{,}900$ images) excluding the recipient's own group. For each question, we shuffle the
pool with a fixed per-question seed, scan at most $64$ candidates, and accept the first two
whose clean margin does not favor the recipient's correct answer ($s\,m(D,q)\le 0$). Groups
are kept only when both questions find two donors, and bootstraps resample groups with both
questions and both donors together.

\begin{figure}[!htbp]
\centering
\includegraphics[width=0.96\linewidth]{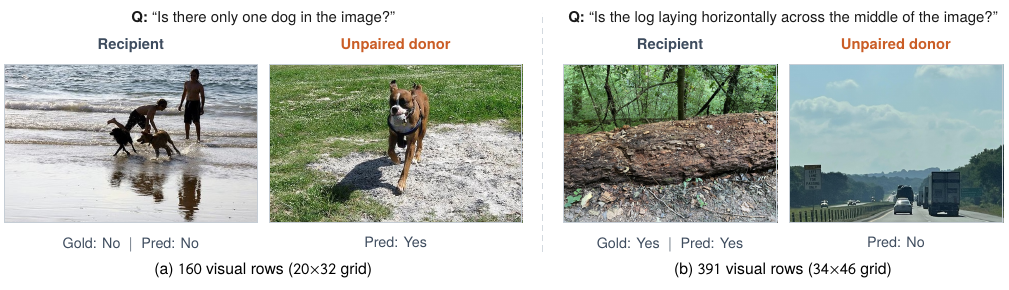}
\caption{Examples of native-resolution unpaired donors used by the Qwen3.5-4B ledger.
Recipients and donors come from different NaturalBench groups and share the exact processor
patch grid without extra padding. Each donor is accepted when its clean margin does not
favor the recipient's ground-truth answer ($s\,m(D,q)\le 0$); donors are not annotated
counterfactuals for the recipient question.}
\label{fig:donor-examples}
\end{figure}

\paragraph{Resolutions and layout.}
At the original resolution, every image is placed on the same $512\times512$ canvas as the
matched ledger, giving $121$ visual rows, so every donor shares the recipient's layout. This
keeps $162$ groups ($648$ recipient--donor pairs). At native resolution, images keep their
aspect ratio within $65{,}536$--$409{,}600$ pixels and donors are drawn only from candidates
with the recipient's exact visual grid. This keeps $153$ groups ($612$ pairs) with $128$--$391$
visual rows (median $192$). In both settings each pair has identical input ids, masks, image
grid, MRoPE positions, and answer-scoring positions (Figure~\ref{fig:donor-examples}).

\begin{figure}[!htbp]
\centering
\includegraphics[width=0.95\linewidth]{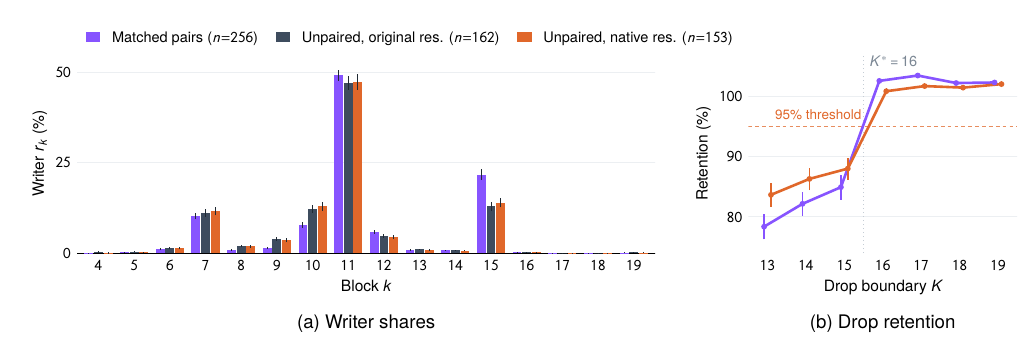}
\caption{Qwen3.5-4B ledgers with unpaired donors. (a) Writer shares with matched pairs and
with unpaired donors at the original and native resolution ($90\%$ intervals); blocks outside
$L4$--$L19$ carry less than $1\%$ in every setting. (b) Drop retention near closure for
matched pairs and native-resolution unpaired donors ($95\%$ intervals). Both are first
certified at $K^*=16$.}
\label{fig:donor}
\end{figure}

\paragraph{Ledgers and eviction.}
Unpaired donors leave the dominant writer in place (intervals in Figure~\ref{fig:donor}a).
$L11$ carries $47.0\%$ at the original and $47.2\%$ at native resolution, versus $49.1\%$
with matched pairs, and the 32-block share profiles correlate with the matched ledger at
Pearson $0.981$ and $0.982$. Only the secondary writers shift: $L15$ falls from $21.7\%$ to
$12.9\%$ and $13.9\%$, while $L10$ rises from $7.7\%$ to $12.2\%$ and $12.9\%$, leaving the
two tied. The two donors of each question give nearly identical profiles (Pearson $0.999$),
and closure remains $K16$ in both settings and for each donor separately. At native
resolution, eviction scanned at $K13$--$K19$ and $K31$ is certified safe from $K16$
($100.8\%$) but not at $K15$ ($88.0\%$), matching the paired-donor boundary (intervals in
Figure~\ref{fig:donor}b).

\suppfloatbarrier
\section{Deployment-scale ledger, boundary scans, and selectors}
\label{app:vl4b}

\paragraph{Ledger calibration.}
The Qwen3-VL-4B-Instruct ledger follows Section~\ref{sec:estimand} on the $256$-group
NaturalBench cohort, and all audits of Supp.~\ref{app:audit} are exactly zero. Cumulative
transfer reaches $29.5\%$ after $L13$, $63.0\%$ after $L15$, $81.9\%$ after $L18$, and
$99.0\%$ after $L19$, where the $90\%$ interval of the remaining fraction enters and stays
inside the $\pm5\%$ band, giving $k^{*}=19$ and $K^{*}=20$. The largest writers are $L15$
($28.6\%$) and $L13$ ($20.9\%$), followed by $L19$ ($17.1\%$) and $L18$ ($10.7\%$; intervals
in Figure~\ref{fig:ledger-vl}a). Because every block is full attention, the main text
protects the top two writers ($L15$ and $L13$) directly. Eviction on the same cohort confirms
the boundary (Supp.~\ref{app:vl-drop}).

\paragraph{Scan grid.}
The scan of Figure~\ref{fig:boundary-scan} serves the deployed checkpoint with visual tokens
evicted at the input of block $K$, and scores fixed stride subsamples (every $10$th POPE,
$6$th EmbSpatial, and $4$th CV-Bench item) with greedy decoding. Qwen3-VL-4B is scanned at
$K\in\{3,6,8,10,\dots,22,24,28,32,35\}$ and Qwen3.5-4B at
$K\in\{2,5,6,8,\dots,18,20,24,28,31\}$, at every block through the recovery region.
Qwen3-VL cannot be scanned below $K3$ because DeepStack injects visual features at
$L0$--$L2$. Each curve's reference is its deepest boundary, which keeps the visual rows
through all but the last block.

\paragraph{Attention-based selectors.}
\label{app:selectors}
The attention rows of Table~\ref{tab:compression} rank blocks by the visual-attention
signals used in FastV and VTW, isolating layer selection from other components of those
methods. Per-block attention mass on visual columns is extracted from an unmodified BF16
eager forward, summed over heads, averaged over $20$ prompts per benchmark, and read either
at the final prompt position (FastV) or averaged over text rows (VTW). Matching the ledger's
protection budget, we protect the top full-attention blocks inside the quantization window:
$L8{+}L15$ (FastV) and $L17{+}L18$ (VTW) on Qwen3-VL. On Qwen3.5 both signals recover the
ledger pair $L11{+}L15$, so the control protects the next two full-attention blocks by writer
share, $L7{+}L3$. On LLaVA-1.5-7B, the mass is instead measured on the $256$-group
NaturalBench cohort at the answer query. Since the answers are a single token, the FastV and
VTW readings coincide and select $L4{+}L8{+}L13$ within the $L3$--$L13$ quantization window.

\paragraph{Single-block protection scan.}
\label{app:kv-scan}
To test whether writer share orders protection benefit across the full compression window, we
protect each of Qwen3-VL-4B's $17$ quantized blocks ($L3$--$L19$) in turn under visual-row A2
quantization with eviction at $K20$, evaluating $889$ A-OKVQA validation items by four-choice
length-normalized likelihood. Writer shares are taken from the $256$-group NaturalBench ledger
above, so no A-OKVQA item enters the ledger or boundary. Each run restores
to BF16 either one block's visual-row $K/V$ inputs or all of its visual-row linear inputs
(whole-block protection), and measures accuracy recovery over unprotected A2 ($37.80\%$; BF16
is $80.99\%$ and eviction alone is $80.09\%$, a $43.19$~pp gap).

\begin{figure}[!htbp]
\centering
\includegraphics[width=0.95\linewidth]{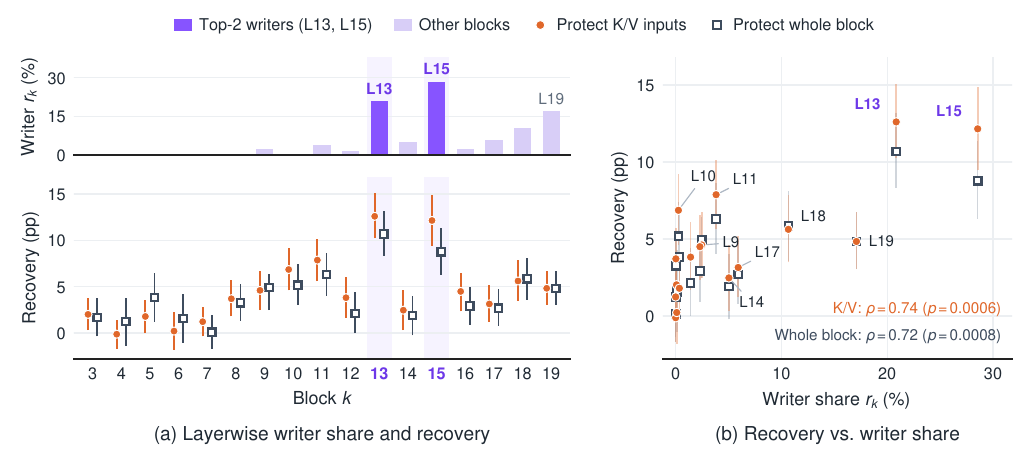}
\caption{Single-block protection on Qwen3-VL-4B (A-OKVQA, $n=889$). \textbf{(a)}~Writer share
$r_k$ (top) and recovery over unprotected A2 when one block's visual-row $K/V$ inputs (circles)
or all visual-row linear inputs (squares) are kept in BF16 (bottom; $95\%$ paired
item-bootstrap CIs). \textbf{(b)}~Recovery against writer share.}
\label{fig:protection-scan}
\end{figure}

Across all $17$ blocks, writer share ranks recovery (Figure~\ref{fig:protection-scan}):
Spearman rank correlation is $0.74$ for $K/V$ protection and $0.72$ for whole-block protection
(one-sided layer-permutation $p<0.001$ for both). The top two ledger writers,
$L13$ ($20.9\%$) and $L15$ ($28.6\%$), yield the two largest recoveries under both scopes
($+12.60$ and $+12.15$~pp for $K/V$; $+10.69$ and $+8.77$~pp for whole-block), whereas none of
the eight blocks with $r_k<2\%$ exceeds $+6.86$~pp ($K/V$) or $+5.17$~pp (whole-block).
Protecting only the $K/V$ inputs matches or exceeds whole-block recovery on $12/17$ blocks,
consistent with $K/V$ carrying the primary visual-to-text pathway.

Local deviations remain: $L11$ ($3.8\%$, $+7.87$~pp $K/V$) and $L10$ ($0.3\%$, $+6.86$~pp)
recover more than neighbors with comparable shares, whereas the third-largest writer $L19$
($17.1\%$) recovers $+4.84$~pp (sixth for $K/V$, seventh for whole-block). Both scopes score
identically at $L19$: eviction at $K20$ discards its visual-row outputs, so text reads $L19$
only through its keys and values.
Task shift between the NaturalBench calibration cohort and A-OKVQA may contribute to these
local shifts; at $n=889$ the $L13$ and $L15$ intervals overlap, and the scan covers a single
checkpoint, task, and bit width.

\suppfloatbarrier
\section{Dose curves under quantization}
\label{app:dose}

We re-run the Qwen3.5-0.8B ledger on the NaturalBench cohort with visual-row activations
quantized over $L3$--$L11$ (all linear inputs except linear-attention projections), measuring
all $24$ boundaries per bit width (Figure~\ref{fig:dose}).

Under A2 (only $12.5\%$ of the clean visual effect retained), $L7$ stays weakly positive
(max-T $p=0.001$) while $L11$ is no longer significant ($p=0.182$); under the shorter
$L3$--$L9$ window, $L11$'s A2 marginal stays at $0.393$, showing that its collapse stems from
extending quantization to $L10$--$L11$. Relative to the surviving effect, closure still holds
after $L11$ in every condition: even under A2, remaining dependence falls inside the $5\%$
band after $L11$ ($1.8\%$, $90\%$ upper bound $3.8\%$).

\input{sections/dose_figure}

\suppfloatbarrier
\section{Protection attribution}
\label{app:ablation}

Table~\ref{tab:protection-attribution} gives the full accuracies, $95\%$ CIs, and
family-wise max-T $p$-values behind Table~\ref{tab:ablation} (A-OKVQA indices
$256$--$3199$; asymmetric per-token A2 with linear-attention projections in BF16; clean BF16
$59.48\%$). Within $L7$, attention protection exceeds MLP by $+6.83$ $[+5.57,+8.12]$~pp and
$K/V$ exceeds both $Q$ and $O$ (all max-T $p<10^{-5}$).

\begin{table}[!htbp]
\centering\small
\caption{Protection targets and paired contrasts on Qwen3.5-0.8B, A-OKVQA ($n=2{,}944$), with
visual-row A2 over $L3$--$L11$ and eviction at $K12$. Recovery is relative to unprotected
A2 ($42.19\%$); pointwise $95\%$ CIs are shown in gray, and $p$-values are family-wise max-T.}
\label{tab:protection-attribution}
\begin{tabular}{lr r@{\,}l r}
\toprule
Target / contrast & Accuracy & \multicolumn{2}{c}{Recovery (pp)} & max-T $p$ \\
\midrule
$L7$ & $49.83\%$ & $+7.64$ & \cisub{$[+6.35,+8.93]$} & $<10^{-5}$ \\
$L3$ & $44.87\%$ & $+2.68$ & \cisub{$[+1.53,+3.84]$} & $4.5\times10^{-5}$ \\
$L4{+}L5$ & $43.75\%$ & $+1.56$ & \cisub{$[+0.65,+2.48]$} & $0.007$ \\
$L8{+}L9$ & $42.63\%$ & $+0.44$ & \cisub{$[-0.17,+1.05]$} & $0.702$ \\
\midrule
\multicolumn{5}{l}{\emph{Same-family contrasts}} \\
$L7-L3$ & & $+4.96$ & \cisub{$[+3.63,+6.35]$} & $<10^{-5}$ \\
$L7-\text{position interpolation}$ & & $+6.78$ & \cisub{$[+5.63,+7.98]$} & $<10^{-5}$ \\
$(L8{+}L9)-(L4{+}L5)$ & & $-1.12$ & \cisub{$[-2.07,-0.20]$} & $0.085$ \\
\midrule
\multicolumn{5}{l}{\emph{Operator and port within $L7$}} \\
attention only & $49.42\%$ & $+7.24$ & \cisub{$[+5.94,+8.53]$} & $<10^{-5}$ \\
MLP only & $42.60\%$ & $+0.41$ & \cisub{$[-0.27,+1.05]$} & $0.675$ \\
$K/V$ only & $49.35\%$ & $+7.17$ & \cisub{$[+5.94,+8.39]$} & $<10^{-5}$ \\
$Q$ only & $42.12\%$ & $-0.07$ & \cisub{$[-0.82,+0.65]$} & $1.000$ \\
$O$ only & $41.58\%$ & $-0.61$ & \cisub{$[-1.29,+0.07]$} & $0.309$ \\
\midrule
\multicolumn{5}{l}{\emph{Second writer's $K/V$ port}} \\
$L11$ $K/V$ only & $45.82\%$ & $+3.64$ & \cisub{$[+2.62,+4.69]$} & $<10^{-5}$ \\
$L7{+}L11$ $K/V$ & $52.00\%$ & $+9.82$ & \cisub{$[+8.32,+11.31]$} & $<10^{-5}$ \\
\bottomrule
\end{tabular}
\end{table}

\suppfloatbarrier
\section{Weight-only protection of the writer blocks}
\label{app:negative}

Under four-bit round-to-nearest weight quantization (W4 RTN) of Qwen3.5-0.8B with eviction
at $K12$, we keep selected self-attention $Q/K/V/O$ weights in BF16 and evaluate $256$
A-OKVQA items (indices $256$--$511$) by four-choice likelihood. The writer pair $L7{+}L11$
is compared with a parameter- and shape-matched control, $L3{+}L15$
(Table~\ref{tab:weight-protection}). No condition yields significant recovery, and the writer
pair does not outperform the control: $\mathrm{Rec}(L7{+}L11)-\mathrm{Rec}(L3{+}L15) = -3.12$ $[-6.25,0.00]$~pp,
max-T $p=0.160$.

\begin{table}[!htbp]
\centering\small
\caption{Weight-only protection on Qwen3.5-0.8B, A-OKVQA ($n=256$), under W4 RTN and eviction
at $K12$. Selected attention weights remain in BF16. Recovery is the accuracy gain
over W4, in pp with pointwise $95\%$ CIs in gray.}
\label{tab:weight-protection}
\begin{tabular}{lr r@{\,}l}
\toprule
Condition & Accuracy & \multicolumn{2}{c}{Recovery (pp)} \\
\midrule
W4 baseline & $52.73\%$ & \multicolumn{2}{c}{---} \\
$L7$ attention BF16 & $53.52\%$ & $+0.78$ & \cisub{$[-1.95,+3.52]$} \\
$L11$ attention BF16 & $54.30\%$ & $+1.56$ & \cisub{$[-0.78,+3.91]$} \\
$L7{+}L11$ attention BF16 & $52.73\%$ & $0.00$ & \cisub{$[-2.73,+2.73]$} \\
$L3{+}L15$ attention BF16 & $55.86\%$ & $+3.12$ & \cisub{$[-0.39,+6.64]$} \\
\bottomrule
\end{tabular}
\end{table}

\suppfloatbarrier
\section{Task-specific transfer and readout mismatch}
\label{app:taskmap}

Capability-level ledgers on MMStar expose a mismatch between prompt-side transfer and
generation. Every capability retains less than $2.4\%$ residual dependence at $K16$, and
mathematics and logical reasoning complete their transfer earlier than perception and instance
reasoning; yet these two capabilities lose $18.0$ and $13.6$~pp under generative eviction at
$K16$, while the other four show no detected change (Supp.~\ref{app:readout}).

\paragraph{Cohort and intervention.}
We run Qwen3.5-4B ledgers on MMStar (Table~\ref{tab:task-ledgers}), splitting each capability
into $16$ pilot, $112$ confirmation, and $122$ holdout items. Inference uses the $669$
eligible confirmation items (of $672$), leaving the holdout split untouched. Each item
receives two same-capability natural donors from the map cohort whose clean margin on the
recipient question is $\le 0$. We score the four-choice length-normalized likelihood margin
on images aspect-preservingly bicubic-padded to $512\times512$ (matching donor and recipient
visual rows one-to-one), evaluating all $33$ boundaries per donor. Bootstraps keep both
donors of an item together.

\begin{table}[!htbp]
\centering\small
\caption{Capability-specific transfer on Qwen3.5-4B, MMStar: $n$ counts the eligible
confirmation items ($669$ total). Writer shares and remaining dependence (at eviction inputs
$K12$ and $K16$, i.e., $\rho_{11}$ and $\rho_{15}$) are percentages of the clean visual effect,
not accuracy; the NaturalBench row is a calibration reference.}
\label{tab:task-ledgers}
\begin{tabular}{lrrrrrrr}
\toprule
& & \multicolumn{4}{c}{Writer share} & \multicolumn{2}{c}{Remaining dep.} \\
\cmidrule(lr){3-6}\cmidrule(lr){7-8}
Capability & $n$ & $L7$ & $L10$ & $L11$ & $L15$ & $\rho_{11}$ & $\rho_{15}$ \\
\midrule
coarse perception & 112 & $9.93\%$ & $15.51\%$ & $38.02\%$ & $17.97\%$ & $25.73\%$ & $1.71\%$ \\
fine-grained perception & 112 & $3.93\%$ & $12.81\%$ & $47.76\%$ & $15.45\%$ & $20.37\%$ & $0.67\%$ \\
instance reasoning & 111 & $13.68\%$ & $12.06\%$ & $42.39\%$ & $17.05\%$ & $25.86\%$ & $0.71\%$ \\
logical reasoning & 112 & $27.11\%$ & $8.89\%$ & $39.83\%$ & $10.24\%$ & $17.45\%$ & $1.80\%$ \\
math & 111 & $23.70\%$ & $10.02\%$ & $34.46\%$ & $7.11\%$ & $14.10\%$ & $1.33\%$ \\
science \& technology & 111 & $19.83\%$ & $13.57\%$ & $32.49\%$ & $8.82\%$ & $16.70\%$ & $2.34\%$ \\
\midrule
confirmation overall & 669 & $15.28\%$ & $12.44\%$ & $39.78\%$ & $14.01\%$ & $21.47\%$ & $1.36\%$ \\
NaturalBench reference & --- & $10.25\%$ & $7.70\%$ & $49.13\%$ & $21.68\%$ & --- & --- \\
\bottomrule
\end{tabular}
\end{table}

Reasoning-heavy capabilities (logical reasoning, mathematics, and science; $n=334$) complete
more of their transfer by $K12$ than perception and instance reasoning ($n=335$), with
remaining dependence lower by $8.2$~pp ($90\%$ CI $[4.0, 12.6]$) and more of their transfer
written by the earlier writer $L7$ (Figure~\ref{fig:taskmap-readout}a). By $K16$ the two
groups show no reliable difference.

\suppfloatbarrier
\subsection{Generative readout under eviction}
\label{app:readout}

\begin{figure}[!htbp]
\centering
\includegraphics[width=\linewidth]{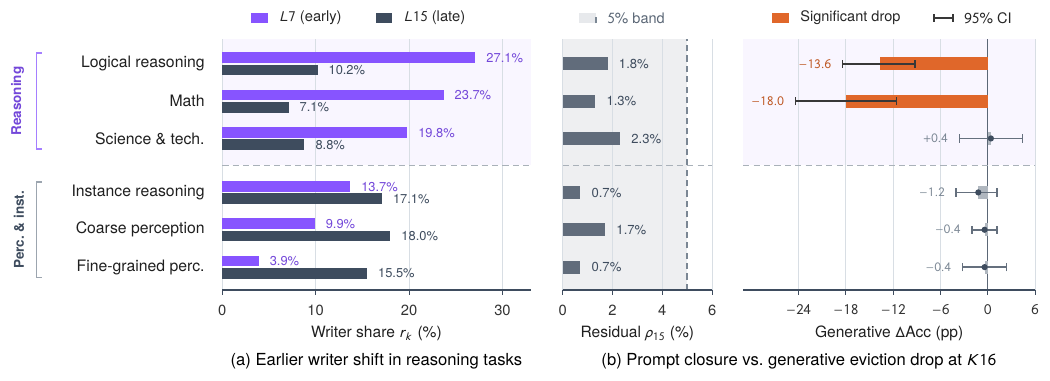}
\caption{Capability-specific transfer and generative eviction on Qwen3.5-4B (MMStar).
(a)~Writer shares for the early ($L7$) and late ($L15$) writers on the confirmation split
(Table~\ref{tab:task-ledgers}). (b)~Prompt-side remaining dependence at $K16$ ($\rho_{15}$,
all well inside the $5\%$ closure band) alongside generative accuracy changes
($\mathrm{Evict}-\mathrm{BF16}$, with pointwise $95\%$ CIs) on all $1{,}500$ items ($250$
per capability).}
\label{fig:taskmap-readout}
\end{figure}

Figure~\ref{fig:taskmap-readout} contrasts the prompt-side transfer profile with standard
generative accuracy across all $1{,}500$ MMStar items ($250$ per capability) under eviction
at $K16$. Overall micro accuracy drops from $66.13\%$ (BF16) to $60.60\%$ ($-5.53$
$[-7.20,-3.87]$~pp).

\suppfloatbarrier
\section{Late-layer visual access during decoding}
\label{app:decode-visual}

\subsection{Conditions and protocol}
\label{app:access-protocol}

\paragraph{Protocol.}
Both models use the same $300$ MMStar items in Table~\ref{tab:access}---$100$ each from
mathematics, logical reasoning, and science \& technology, sampled unfiltered from the
non-holdout task-map cohort---with BF16 greedy decoding ($1{,}024$-token budget) and the
original lmms-eval parser. For Block reads, both arms share the clean prefill and first output
token before diverging; the blocked arm masks attention from generated queries to visual keys
while keeping visual $K/V$ and all block computations intact. Qwen3.5-4B uses the original
prompt in non-thinking mode and blocks its post-closure full-attention layers
($L19/L23/L27/L31$); Qwen3-VL-4B-Instruct blocks $L20$--$L35$ from $K20$ and omits only the
trailing instruction ``Answer with the option letter only.''

\paragraph{Estimation and supplementary results.}
Bootstraps are stratified by capability and device. Pooled across the three capabilities,
blocking shifts accuracy by $-10.33$ $[-14.67,-6.00]$~pp on Qwen3.5 and $-8.33$
$[-13.33,-3.00]$~pp on Qwen3-VL. Keeping the letter-only instruction leaves $291/300$
Qwen3-VL outputs as a single clean token ($26/300$ without it; $-1.00$~pp change under
blocking); similarly, the shared first token already outputs an option on $85/100$ Qwen3.5
science items. Unparsed responses count as incorrect and token-capped items are retained;
correct-to-incorrect flips also occur when neither arm hits the cap.

\paragraph{Eviction and freezing.}
Evict and Freeze share this evaluation setup but intervene during prefill, so later prompt
states and the first output token can change. Evict drops all visual rows at block input $K$
($K16$ on Qwen3.5-4B, $K20$ on Qwen3-VL-4B). Freeze instead fixes visual rows to block
$K{-}1$'s clean output: each later full-attention block applies its own normalization and
$K/V$ projections (preserving original MRoPE positions) into its own unshared cache during
prefill, so generated tokens read them via standard attention. On Qwen3-VL-4B ($36$
full-attention blocks), visual inputs reset to these frozen states from $K$ onward; on
Qwen3.5-4B, GDN blocks from $K$ onward scan text rows alone (evolving their recurrent state on
text only), while each later full-attention block ($L19/L23/L27/L31$ for $K16$) reinserts the
frozen visual rows at their original positions and keeps only its text outputs.

Pooled recovery of eviction loss,
$(a_{\mathrm{Freeze}}-a_{\mathrm{Evict}})/(a_{\mathrm{Full}}-a_{\mathrm{Evict}})$, is $30/31$
lost answers on Qwen3.5-4B and $33/38$ on Qwen3-VL-4B. The $\approx 5\%$ cost in
Section~\ref{sec:limits} is the analytic per-token linear MAC ratio of projecting a frozen row
through $K/V$ alone versus a full SwiGLU block ($Q/K/V/O$ and $3$-matrix MLP; hidden width $d$
cancels): $2d_{kv}/(2d_q{+}2d_{kv}{+}3d_m)$ (with $3d_q$ for Qwen3.5's gated query projection),
giving $2{,}048/39{,}424=5.19\%$ on Qwen3-VL-4B ($d_q{=}4{,}096, d_{kv}{=}1{,}024,
d_m{=}9{,}728$) and $2{,}048/41{,}984=4.88\%$ per full-attention block on Qwen3.5-4B
($d_m{=}9{,}216$; lower when averaged with $3{:}1$ text-only GDN blocks). Accuracy runs
evaluate the equivalent frozen states without this kernel optimization, and
Figure~\ref{fig:freeze} sweeps $K$ on mathematics and logical reasoning only (science is
unaffected at closure).

\subsection{Reading facts from the prompt}
\label{app:fact-probe}

\begin{figure}[!htbp]
\centering
\includegraphics[width=\linewidth]{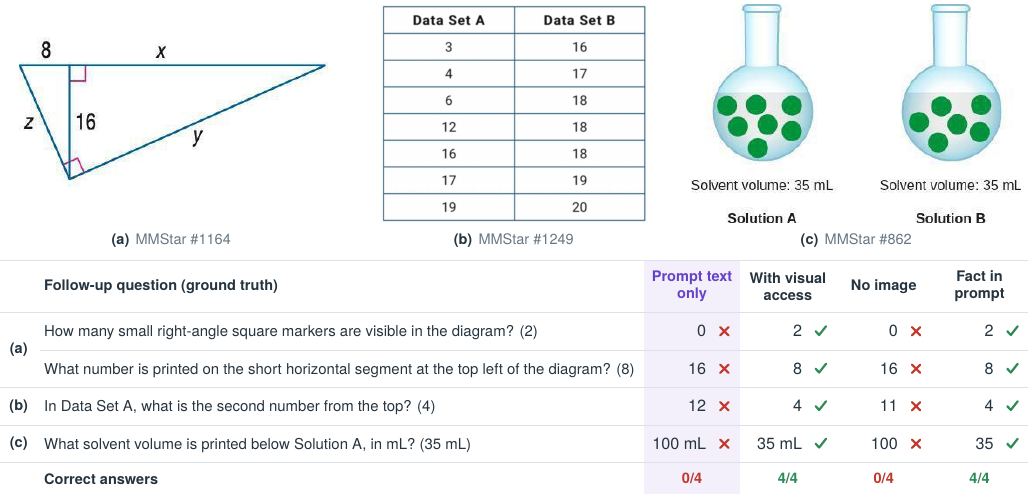}
\caption{Visual-fact follow-ups on Qwen3-VL-4B after the clean prefill of three MMStar items.
Ground-truth answers are in parentheses and responses are reproduced verbatim. \emph{Prompt
text only} blocks the follow-up from the visual keys, and \emph{With visual access} does not.
\emph{No image} and \emph{Fact in prompt} repeat the blocked follow-up after a prefill without
the image and with the fact stated in the prompt.}
\label{fig:fact-cases}
\end{figure}

\paragraph{Protocol.}
Figure~\ref{fig:fact-cases} probes four visual facts across three MMStar items on Qwen3-VL-4B
(Section~\ref{sec:limits}). After clean prefill of the original image--question prompt
without generating an answer, we append a forced continuation---an assistant cue (``I will
check one visual fact.''), a user probe requesting a short factual answer only, and an open
assistant turn decoded greedily (up to $128$ tokens, never reached)---and score responses
against normalized ground-truth aliases. Starting from identical prefill caches, \emph{Prompt
text only} masks attention from all appended and generated tokens to visual keys across all
$36$ blocks while leaving non-visual prompt states intact, whereas \emph{With visual access}
leaves attention unmodified; \emph{No image} and \emph{Fact in prompt} repeat the blocked
follow-up from fresh prefills without the image or with the answer stated in one appended
sentence.

\makeatletter\setlength{\@fptop}{0pt}\makeatother

\paragraph{Scope.}
Because \emph{With visual access} and \emph{Fact in prompt} both score $4/4$ while
\emph{Prompt text only} scores $0/4$ (repeating the no-image answer on three of the four probes), the
failures reflect missing readable information in the prompt's text states rather than a
masking or forced-turn artifact. Only the marker count marks the exact divergence point
between Full and Block reads; the remaining three probes test additional image details (and
blocking does not hurt item~(c)). As four diagnostic cases rather than a benchmark estimate,
failed readouts establish only that this follow-up probe cannot retrieve the facts from
retained non-visual states (including image-boundary tokens), not that prefill failed to
encode them.

%% file: sections/edge_scan_figure.tex
\begin{figure}[!htbp]
\centering
\includegraphics[width=0.92\linewidth]{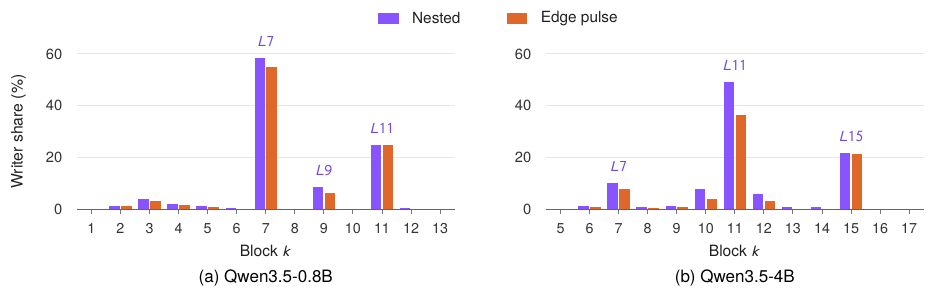}
\caption{Nested and edge-pulse writer shares across active blocks ($L1$--$L13$ on
Qwen3.5-0.8B; $L5$--$L17$ on Qwen3.5-4B). Both schedules are normalized by the nested run's
mean clean total effect; every omitted block carries less than $0.4\%$ in both schedules.}
\label{fig:edge-scan}
\end{figure}

%% file: sections/drop_scan_extrema.tex
Across all $K < K^*$, retention upper bounds never exceed $70.7\%$ ($0.8$B) and $87.0\%$ ($4$B), whereas lower bounds from $K^*$ onward remain at least $99.8\%$ and $99.6\%$.

%% file: sections/drop_scan_figure.tex
\begin{figure}[!htbp]
\centering
\includegraphics[width=0.95\linewidth]{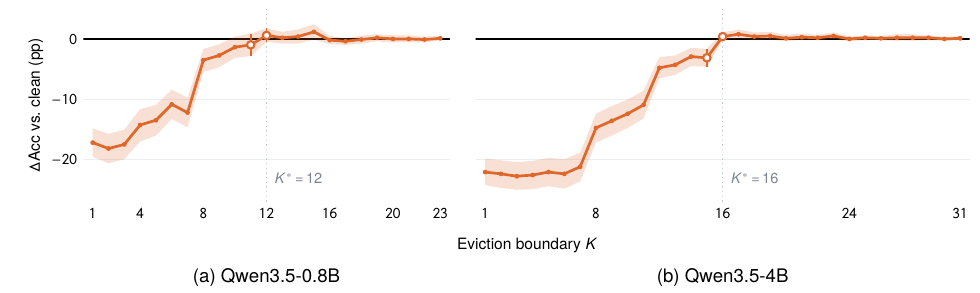}
\caption{Accuracy change versus clean at every eviction boundary on NaturalBench. Ribbons show pointwise $95\%$ intervals; open circles mark $K^*{-}1$ and $K^*$, and dotted lines mark $K^*$. Safety is certified by clean-gap retention, not accuracy.}
\label{fig:drop-scan}
\end{figure}

%% file: sections/dose_figure.tex
\begin{figure}[!htbp]
\centering
\includegraphics[width=0.82\linewidth]{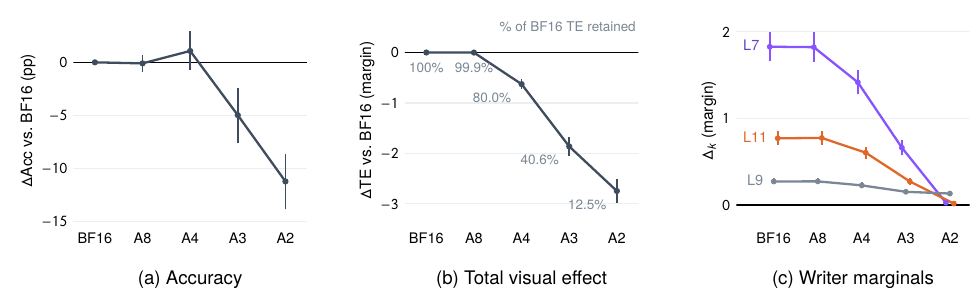}
\caption{Dose curves under $L3$--$L11$ visual-row quantization (Qwen3.5-0.8B, NaturalBench). (a)~Accuracy and (b)~total-effect change versus BF16; labels in (b) give the fraction of the BF16 effect retained. (c)~Ordered marginals of the three BF16 writers. Whiskers are pointwise $95\%$ group-bootstrap CIs.}
\label{fig:dose}
\end{figure}